\documentclass[11pt]{article}

\usepackage[final]{acl}

\usepackage{times}
\usepackage{latexsym}
\usepackage{amsmath}
\usepackage{enumerate}
\usepackage{enumitem}
\usepackage{amssymb}
\usepackage{booktabs}
\usepackage{supertabular}
\usepackage{multirow}
\usepackage{makecell}
\usepackage{algorithm}
\usepackage{algorithmic}
\usepackage{longtable}
\usepackage{array}
\usepackage{balance}
\usepackage{hyperref}
\usepackage[most]{tcolorbox}
\newtcolorbox{researchquestionbox}{
  enhanced,
  colback=gray!5,
  colframe=gray!45,
  boxrule=0.6pt,
  arc=2pt,
  left=6pt,
  right=6pt,
  top=5pt,
  bottom=5pt
}
\usepackage[T1]{fontenc}
\usepackage[utf8]{inputenc}

\usepackage{microtype}

\usepackage{inconsolata}

\usepackage{graphicx}
\newcommand{\std}[1]{{\scriptsize$\pm$#1}}

\title{Towards Automatic Evolution Tree Generation from Citation Graphs}

\author{
  \textbf{Zexing Zhao}\textsuperscript{1}, \quad
  \textbf{Yuntong Hu}\textsuperscript{2}, \quad
  \textbf{Liang Zhao}\textsuperscript{2\textdagger} \\
  \textsuperscript{1}School of Mechanical Engineering, Georgia Institute of Technology \\
  \textsuperscript{2}Department of Computer Science, Emory University \\
  \texttt{alfred.zhao@gatech.edu}, \quad
  \texttt{\{yuntong.hu, liang.zhao\}@emory.edu}
}
\begin{document}
\maketitle

\begin{abstract}
Surveys remain the primary way researchers grasp the lineage of methods within an AI subfield, but they scale poorly against the current rate of publication. Existing taxonomy-induction methods are largely leaf-bound and time-agnostic; they tend to force transitional papers into mature leaves and can create topological inversions between ancestors and descendants. We propose EvoTree, a staged framework that decouples conceptual backbone learning from temporal refinement: a graph-aware encoder with distribution-based hierarchical clustering yields a stable taxonomy backbone; temporal fine-tuning then re-attaches marginal papers to internal nodes under monotonic-path constraints; a final LLM pass labels concepts without altering the topology. We release the first annotated benchmark for this task across 11 AI subfields. EvoTree attains the highest NMI and citation-direction accuracy among all baselines and the best concept purity on the annotated benchmark, and is the only method with non-trivial marginal-paper detection on the annotated set.
\end{abstract}
\blfootnote{\textsuperscript{\textdagger}Corresponding author.}

\section{Introduction}
The rapid growth of scientific literature has fundamentally challenged human-centered knowledge organization and field understanding. Open scholarly infrastructures such as OpenAlex~\cite{priem2022openalex} now index over 470 million scientific works together with continuously expanding citation networks. As research fields evolve at an unprecedented pace, researchers increasingly struggle to efficiently understand the historical evolution, conceptual lineage, and paradigm transitions of rapidly developing domains~\cite{fortunato2018science}, creating an increasing need for AI-assisted methods to organize, synthesize, and understand scientific progress~\cite{le2026evonarrator, hu2025cg}.

Among various forms of scientific organization, evolution trees and hierarchical lineage diagrams are particularly effective because they explicitly reveal conceptual inheritance, branching, and progressive refinement across generations of work. Such structures are widely adopted in modern survey papers to summarize the evolution of research paradigms, model families, and emerging subfields~\cite{zhang2018taxogen,wang2024auto,hu2024hi,liang2025survey}. However, constructing these evolution trees remains largely manual, requiring substantial domain expertise and intensive literature analysis, while quickly becoming outdated as new papers continuously emerge~\cite{wang2024auto,liang2025survey}. This motivates a fundamental research question:

\begin{researchquestionbox}
\noindent\textit{\textbf{Can we automatically generate scientific evolution trees from citation graphs?}}
\end{researchquestionbox}

Addressing this problem is highly challenging due to several underexplored issues. First, scientific evolution trees are hierarchical, temporal, and semantically structured objects, whereas citation graphs are large-scale sparse relational networks. Bridging these two fundamentally different structures requires learning evolution-aware representations that preserve both citation dependencies and temporal scientific progression, which is not directly addressed by existing taxonomy induction methods~\cite{zhang2018taxogen,shen2020taxoexpan,yu2020steam} or temporal topic modeling approaches~\cite{blei2006dynamic,wang2006topics,wang2008continuous}. Second, scientific evolution is inherently dynamic: research paradigms continuously branch, merge, and refine over time, making it difficult to construct stable and interpretable lineage structures from evolving citation graphs. Existing citation and co-citation analyses in science-of-science research primarily characterize global scientific trends rather than inducing explicit hierarchical evolutionary structures~\cite{fortunato2018science}. Third, existing survey papers provide only limited supervision signals, requiring models to generalize from scarce high-quality evolution annotations. Moreover, marginal or transitional papers often lie between major paradigms, making them difficult to position within conventional taxonomy induction frameworks without introducing chronological inconsistencies.

% Crucially, identifying marginal papers is not a problem the clustering stage alone can solve. Without temporal evidence, an "atypical" paper is geometrically indistinguishable from an outlier; only when chronological context is available can one tell whether a paper is a transitional ancestor or a noisy outlier. This observation motivates a staged design in which marginal-node modeling is deliberately deferred until temporal signals become available.

To address those challenges, we propose a structurally staged and temporally constrained framework that unifies taxonomy induction, temporal admissibility, and marginal-paper re-attachment into a cohesive pipeline. Specifically, our framework first constructs a stable, time-agnostic core taxonomy via distribution-based hierarchical clustering on graph-aware node embeddings. Subsequently, it performs temporal refinement to model evolutionary dynamics and handle marginal papers by attaching them to broader internal concepts, while enforcing temporal consistency and structural legality constraints (e.g., path monotonicity). This staged approach ensures that the conceptual backbone remains robust while temporal dynamics are accurately and logically mapped. 

Our contributions are summarized as follows:
\begin{itemize}[leftmargin=*,nosep]
\item \textbf{Novel Problem Formulation.} We formulate evolutionary tree construction from citation graphs, introducing a dynamic perspective for scientific literature mining.  
\item \textbf{Staged \& Constrained Framework.} We propose a staged framework that separates taxonomy initialization from temporal refinement, preserving citation-aware structures while filtering peripheral papers from core trajectories.  
\item \textbf{Evolutionary Dynamics Modeling.} We incorporate temporal and structural constraints with marginal-paper detection and internal-node reattachment, alleviating chronological distortions in traditional taxonomy generation.  
\item \textbf{Comprehensive Benchmark.} We construct the first benchmark for this task across 11 AI subfields, where extensive experiments show consistent gains over taxonomy, citation-based, and LLM baselines in evolution-structure quality and annotated concept purity, together with the only non-trivial marginal-paper detection. 
\end{itemize}

\section{Related Work}

\subsection{Automatic Taxonomy Induction}
Automatic taxonomy induction studies how to organize terms, entities, or topics into hierarchical structures such as trees or DAGs. Prior work has shown that taxonomy quality requires more than isolated pairwise hypernym prediction~\cite{snow2006semantic,bansal2014structured,mao2018end}. Most existing approaches focus on taxonomy expansion or completion given an existing hierarchy~\cite{shen2020taxoexpan,yu2020steam,mishra2025rank,jiang2022taxoenrich}, or extend the hierarchy formalism itself~\cite{lee2022taxocom,lu2024self,kar2025taxoadapt}. A smaller line of work constructs topic or concept hierarchies directly from corpora. TaxoGen~\cite{zhang2018taxogen} recursively builds topic taxonomies through adaptive spherical clustering and local embeddings, while HiExpan~\cite{shen2018hiexpan} expands user-guided seed hierarchies using weakly supervised relation extraction. A recent line of work applies language models to scholarly taxonomy construction. TaxoAlign~\cite{lahiri2025taxoalign} generates taxonomies for scientific domains by aligning LM-induced hierarchies to reference structures, and \citet{zhu-etal-2025-context} induce hierarchical taxonomies for scientific papers via LLM-guided multi-aspect clustering. 
However, these methods are primarily designed for text corpora and cannot effectively model the structural dependencies. Moreover, they generally assume static taxonomies and therefore fail to capture temporally constrained scientific evolution, including paradigm transitions and lineage progression across generations of work.

\subsection{Knowledge Evolution Modeling}
Knowledge evolution modeling studies how scientific concepts, research topics, and paradigms emerge, evolve, and transition over time. Early work primarily focused on temporal topic modeling, including Dynamic Topic Models~\cite{blei2006dynamic,wang2008continuous,2016bhaduryscaling} and Topics over Time~\cite{wang2006topics}, which capture topic drift and temporal word distributions across document collections. Another line of research investigates scientific evolution through citation and co-citation analysis within the science-of-science community~\cite{fortunato2018science}. More recently, LLM-based survey systems such as AutoSurvey~\cite{wang2024auto} and SurveyX~\cite{liang2025survey} retrieve papers, construct outlines, and generate survey-style text, but primarily optimize textual synthesis rather than explicit structural evolution modeling. Although several recent works summarize papers into taxonomy-like structures~\cite{hu2024hi,zhu-etal-2025-context}, none directly addresses scientific evolution tree generation. Existing methods lack temporal constraints over evolutionary paths, and mechanisms for handling marginal papers or constructing survey-style evolution structures. 
% EvoTree targets this missing intersection by inducing temporally coherent scientific lineages from citation graphs.

\section{Problem Formulation}
\label{sec:problem}

\paragraph{Citation graph.} Given a survey paper, let 
$\mathcal{G}=(\mathcal{V},\mathcal{E}_{\mathcal{G}})$ denote the citation graph constructed from its reference set, where each node $v_i\in\mathcal{V}$ is a referenced paper with text content $s_i$ (title and abstract) and publication year $\tau(v_i)$. A directed edge $(v_i,v_j)\in\mathcal{E}_{\mathcal{G}}$ indicates that paper $v_i$ cites paper $v_j$. For the main evaluation, the target survey itself is excluded from $\mathcal{G}$. 

\paragraph{Taxonomy tree.}
A taxonomy tree organizes papers by conceptual inclusion. We denote it as
$\mathcal{T}^{\mathrm{tax}}=(\mathcal{C}^{\mathrm{tax}},\mathcal{V},\mathcal{E}^{\mathrm{tax}},\phi^{\mathrm{tax}})$,
where $\mathcal{C}^{\mathrm{tax}}$ is the set of topic nodes, $\mathcal{V}$ is the set of paper nodes, and
$\mathcal{E}^{\mathrm{tax}}\subseteq \mathcal{C}^{\mathrm{tax}}\times(\mathcal{C}^{\mathrm{tax}}\cup\mathcal{V})$ forms a rooted tree. Each topic node $c\in\mathcal{C}^{\mathrm{tax}}$ represents a semantic concept with label $\phi^{\mathrm{tax}}(c)$, and each paper node $v_i\in\mathcal{V}$ appears only as a terminal leaf (i.e., for every paper node, its parent must be a concept node rather than another paper node, and this parent corresponds to the lowest-level concept cluster that directly contains the paper). Thus, edges in $\mathcal{T}^{\mathrm{tax}}$ encode conceptual refinement only, with no temporal semantics.

\paragraph{Evolutionary tree.}
An evolutionary tree $\mathcal{T} = (\mathcal{C}, \mathcal{V}, \mathcal{E}_C, \mathcal{E}_P, \phi)$ extends this organization with temporally admissible development relations. $\mathcal{E}_C$ forms a rooted tree over concepts $\mathcal{C}$ with labels $\phi$, and $\mathcal{E}_P = \{(a(v_i), v_i)\}$ attaches each paper to a concept via the attachment map $a: \mathcal{V} \to \mathcal{C}$. Unlike a taxonomy tree, papers may attach to internal concept nodes when better explained at a coarser abstraction level.

For each concept $c$, let $\mathcal{V}_{\text{att}}(c) = \{v_i : a(v_i) = c\}$ denote the papers attached directly to $c$, and $\mathcal{V}(c)$ the papers in its full subtree. We define two temporal statistics over publication years $\tau(\cdot)$:
\begin{equation}
\small
\tau^{\text{att}}(c) = \tfrac{1}{|\mathcal{V}_{\text{att}}(c)|}\!\!\!\sum_{v \in \mathcal{V}_{\text{att}}(c)}\!\!\!\tau(v),
\;
\bar{\tau}(c) = \tfrac{1}{|\mathcal{V}(c)|}\!\sum_{v \in \mathcal{V}(c)}\!\tau(v).
\label{eq:tau-defs}
\end{equation}
$\tau^{\text{att}}$ supports structural constraints: because the attachment sets of a parent and its children are disjoint, the constraint it induces is non-trivial. $\bar{\tau}$ summarizes the full concept body and is used for path-level evaluation.

The tree must satisfy (i) \textbf{Rooted Tree}; (ii) \textbf{Temporal Consistency}, for each concept edge $(c_p, c_c) \in \mathcal{E}_C$ from parent $c_p$ to child $c_c$,
\begin{equation}
\tau^{\text{att}}(c_p) \leq \tau^{\text{att}}(c_c) + \epsilon;
\label{eq:temporal-consistency}
\end{equation}
and (iii) \textbf{Sibling Ordering} by non-decreasing $\tau^{\text{att}}$. Eq.~\ref{eq:temporal-consistency} requires each root-to-leaf path to proceed from earlier, broader concepts to later, more specialized ones, with the slack $\epsilon$ absorbing noise in publication dates so that a single mis-dated paper does not invalidate an otherwise valid edge.

% \paragraph{Marginal papers.}
% A marginal paper is a paper that cannot be confidently assigned to any fine-grained evolutionary leaf because it has weak semantic, structural, or temporal compatibility with all leaf-level concepts. Formally, given a continuous compatibility score $m(v_i, c)\in[0,1]$ between paper $v_i$ and concept node $c$ (which will be instantiated as a multi-view score in \S\ref{sec:3.4.3}), we define the marginal set as
% \begin{equation}
%     \mathcal{M}
%     =
%     \left\{
%     v_i\in\mathcal{V}:
%     \max_{\ell\in \mathrm{Leaves}(\mathcal{T})}
%     m(v_i,\ell) < \eta
%     \right\},
% \end{equation}
% where $\eta$ is a threshold. Marginal papers are not discarded; instead, they may be attached to internal concept nodes, allowing the tree to represent isolated, transitional, or weakly continued methodological contributions.

\paragraph{Marginal papers.}
A marginal paper has weak compatibility with all leaf-level concepts. Given a continuous compatibility score $m(v_i, c)\in[0,1]$ (instantiated in \S\ref{sec:3.4.3}), the marginal set is
\begin{equation}
\mathcal{M} = \{v_i \in \mathcal{V} : \max_{\ell\in \mathrm{Leaves}(\mathcal{T})} m(v_i,\ell) < \eta\},
\label{eq:marginal-def}
\end{equation}

where $\eta$ is a threshold. Marginal papers are not discarded; they may attach to internal nodes to represent isolated, transitional, or weakly continued contributions.

\paragraph{Challenges and design overview.}
This formalism poses four challenges that shape our method.
(C1) The discrete structure $(\mathcal{C}, \mathcal{E}_C, a)$ and continuous embeddings $\{\mathbf{e}_i\}$ cannot be jointly optimized end-to-end, motivating \emph{alternating refinement} between structure and representation (\S\ref{sec:3.4.1}).
(C2) Temporal consistency (Eq.~\ref{eq:temporal-consistency}) cannot emerge from a time-agnostic clustering loss, motivating a \emph{temporal merge filter} on $\tau^{\text{att}}$ (\S\ref{sec:3.4.1}).
(C3) Marginal papers fall outside leaf-only partitions, motivating explicit \emph{internal-node re-attachment} (\S\ref{sec:3.4.3}).
(C4) Temporal refinement risks collapsing the conceptual organization, motivating a \emph{staged design} that establishes a stable backbone $\mathcal{T}^{\mathrm{tax}}$ (\S\ref{sec:3.3}) before applying temporal pressure (\S\ref{sec:3.4}).

%\paragraph{Task.}
%Given a citation graph $\mathcal{G}$ built from the reference set of a target survey, together with each paper's title, abstract, and publication year, our task is to generate an evolutionary tree $\mathcal{T}$ that describes the development of methods in the surveyed field. The tree should construct concept-level topology, attach papers to suitable concept nodes, identify marginal papers that cannot be confidently assigned to fine-grained leaves, and produce interpretable concept labels for the resulting nodes.

\section{Methodology}

\begin{figure*}[t]
  \centering
  \includegraphics[width=1\linewidth]{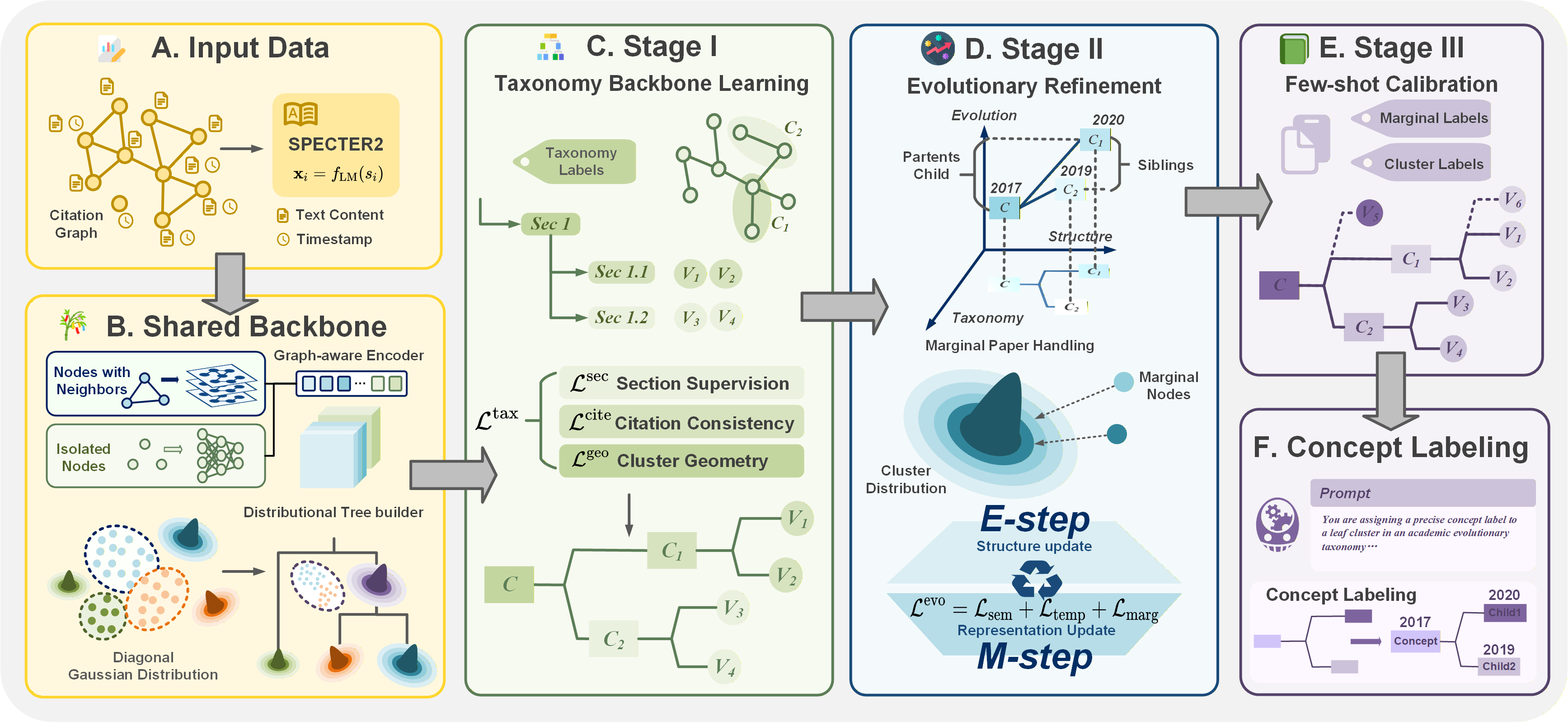}
\caption{Overview of the EvoTree pipeline.
\textbf{(A)} Input citation graph encoded by SPECTER2.
\textbf{(B)} Shared backbone with graph-aware encoder and distributional tree builder.
\textbf{(C)} Stage~I taxonomy backbone learning under $\mathcal{L}^{\text{tax}}$.
\textbf{(D)} Stage~II evolutionary refinement via alternating updates under $\mathcal{L}^{\text{evo}}$.
\textbf{(E)} Stage~III few-shot calibration.
\textbf{(F)} LLM concept labeling.}\label{fig:framework}
\end{figure*}

\subsection{Overview}
\label{sec:overview}
EvoTree (Fig.~\ref{fig:framework}) addresses challenges C1--C4 through three stages over a shared backbone. The input of EvoTree (Fig.~\ref{fig:framework}A) is a citation graph in which each node carries paper text and a timestamp, encoded by SPECTER2. The backbone (\S\ref{sec:3.2}) couples a graph-aware encoder with a distributional tree builder that summarizes each cluster as a diagonal Gaussian. Stage~I (\S\ref{sec:3.3}) learns a time-agnostic taxonomy backbone $\mathcal{T}^{\mathrm{tax}}$ under section-derived weak supervision. Stage~II (\S\ref{sec:3.4}) refines $\mathcal{T}^{\mathrm{tax}}$ into the evolutionary tree $\mathcal{T}$ through alternating structure--representation updates with temporal admissibility and marginal-paper handling. Stage~III (\S\ref{sec:3.5}) calibrates against reference trees via few-shot metric learning, and a final LLM pass (\S\ref{sec:3.6}) verbalizes concept labels without altering topology. The main text is self-contained for the design rationale of each component; appendices provide derivations, loss formulations, and implementation details.

\subsection{Shared Backbone}\label{sec:3.2}

The pipeline rests on two reusable modules: a \emph{graph-aware paper encoder} and a \emph{distributional tree builder}.

\subsubsection{Graph-aware Encoder}
\label{sec:3.2.1}

The encoder maps each paper $v_i$ to a clustering embedding $\mathbf{e}_i \in \mathbb{R}^{d}$ from text and citation context. A pretrained scientific encoder SPECTER2~\cite{Singh2022SciRepEvalAM} produces a time-agnostic vector $\mathbf{x}_i$, from which a semantic stream $\mathbf{z}_i^{\mathrm{sem}}=\mathrm{MLP}_{\mathrm{sem}}(\mathbf{x}_i)$ and a graph stream $\mathbf{h}_i^{(0)}=\mathrm{MLP}_{\mathrm{graph}}(\mathbf{x}_i)$ are derived. The graph stream propagates through $L$ layers with anchored residuals and PairNorm~\cite{zhao2020pairnorm}:
\begin{equation}
\small
\begin{aligned}
\mathbf{h}_i^{(l+1)} = \\\mathrm{PairNorm}&\!\left(\alpha \mathbf{x}_i + (1-\alpha)\big[\beta \mathbf{h}_i^{(l)} + (1-\beta)\mathbf{m}_i^{(l)}\big]\right),
\end{aligned}
\label{eq:graph-prop}
\end{equation}
where $\mathbf{m}_i^{(l)}$ aggregates messages from citation neighbors (set to $\mathbf{0}$ for isolated nodes). The anchored term mitigates over-smoothing while PairNorm stabilizes propagation under uneven density. The two streams fuse via a connectivity-biased gate:
\begin{equation}
\begin{aligned}
    \mathbf{z}_i^{\mathrm{fused}} = \lambda_i \mathbf{h}_i^{(L)} + (1-\lambda_i)\mathbf{z}_i^{\mathrm{sem}},\\
\quad \lambda_i \!=\! \sigma\!\big(\mathbf{W}_g[\cdot] + \delta\,\mathbb{1}[\mathrm{iso}(v_i)]\big),
\end{aligned}
\label{eq:fusion-gate}
\end{equation}
with $\delta<0$ biasing isolated nodes toward the semantic stream. The final embedding is $\mathbf{e}_i=\mathrm{L_2Norm}(\mathbf{z}_i^{\mathrm{fused}})$; a separate projection $\mathbf{p}_i$ is reserved for contrastive learning~\cite{chen2020simple}. Full details in Appendix~\ref{app:encoder}.

\subsubsection{Distributional Tree Builder}
\label{sec:3.2.2}

Given embeddings $\{\mathbf{e}_i\}$, the builder constructs a hierarchical tree through distributional clustering. HDBSCAN with soft membership~\cite{campello2013density} yields base clusters $\{C_k\}_{k=1}^{K}$ and weights $w_{ik}$, each summarized as $C_k \sim \mathcal{N}(\boldsymbol{\mu}_k, \mathrm{diag}(\boldsymbol{\sigma}_k^2))$. Higher-level concepts arise from bottom-up agglomeration under the closed-form 2-Wasserstein distance:
\begin{equation}
d^2(C_p,C_q) = \|\boldsymbol{\mu}_p-\boldsymbol{\mu}_q\|_2^2 + \|\boldsymbol{\sigma}_p-\boldsymbol{\sigma}_q\|_2^2.
\label{eq:w2-dist}
\end{equation}
The parent Gaussian is re-estimated from union members after each merge, reducing level-wise drift. The same builder is used in Stages~I and~II; what changes between them is the supervision and whether temporal admissibility is enforced.

\subsection{Stage I: Taxonomy Backbone Learning}
\label{sec:3.3}

Stage~I learns a time-agnostic backbone $\mathcal{T}^{\mathrm{tax}}$ that captures conceptual inclusion. Inputs are the citation graph, paper texts, and weak hierarchical labels derived from the survey section structure (each paper inherits its section path). We pre-train the graph-aware paper encoder (\S\ref{sec:3.2.1}) under three complementary signals: \textbf{section supervision} (multi-label classification and level prediction over the section hierarchy), \textbf{citation consistency} (link prediction and contrastive ranking on citation edges), and \textbf{cluster geometry} (centroid pulling for labeled papers):
\begin{equation}
\mathcal{L}^{\mathrm{tax}} = \mathcal{L}_{\mathrm{sec}} + \mathcal{L}_{\mathrm{cite}} + \mathcal{L}_{\mathrm{geo}}.
\label{eq:tax-loss}
\end{equation}
Applying the distributional tree builder to the pre-trained embeddings yields $\mathcal{T}^{\mathrm{tax}}$, which serves as the structural prior for Stage~II. Component formulations are given in Appendix~\ref{app:pretrain_loss}.

We emphasize that section structures serve as weak supervision rather than ground-truth concept hierarchies: headings reflect expository choices as much as conceptual organization, and their granularity varies across surveys. Their influence is correspondingly bounded. They contribute only one of the three terms in Eq.~\ref{eq:tax-loss}, shaping the initialization jointly with the citation and geometry objectives rather than acting as a direct optimization target, and are unavailable at inference.

\subsection{Stage II: Evolutionary Refinement}\label{sec:3.4}

Stage~II transforms $\mathcal{T}^{\mathrm{tax}}$ into the evolutionary tree $\mathcal{T}$, driven by three signals absent from Stage~I: \textbf{temporal admissibility} (Eq.~\ref{eq:temporal-consistency}), \textbf{citation directionality} (citing papers attach deeper than cited ones), and \textbf{marginal-paper re-attachment}. The heavy components (SPECTER2, GNN weights) remain frozen; only lightweight adapters (projection MLP, fusion gate, marginal head, cluster Gaussians) are fine-tuned.

\subsubsection{EM-style Refinement}
\label{sec:3.4.1}

Because discrete tree construction and continuous adapter updates cannot be jointly optimized end-to-end (C1), we adopt an alternating procedure inspired by EM. We label the two phases as E-step and M-step for brevity, without claiming a strict EM optimization of a probabilistic likelihood.
 
\paragraph{E-step --- structure update.}
With adapters fixed, we re-apply the distributional tree builder to the current embeddings. Unlike Stage~I, agglomeration is filtered by temporal admissibility: a candidate merge producing parent $c_p$ from children $c_a, c_b$ is rejected if
\begin{equation}
\tau^{\text{att}}(c_p) > \min\!\big(\tau^{\text{att}}(c_a),\, \tau^{\text{att}}(c_b)\big) + \epsilon,
\label{eq:merge-filter}
\end{equation}
which uses the same $\tau^{\text{att}}$ statistic as the formal admissibility condition (Eq.~\ref{eq:temporal-consistency}). This step yields an updated tree $\mathcal{T}^{(t)}$ and a set of low-membership marginal candidates.
 
\paragraph{M-step --- representation update.}
With $\mathcal{T}^{(t)}$ fixed, we update the adapters under three self-supervised objectives:
\begin{equation}
\mathcal{L}^{\mathrm{evo}} = \mathcal{L}_{\mathrm{sem}} + \mathcal{L}_{\mathrm{temp}} + \mathcal{L}_{\mathrm{marg}},
\label{eq:evo-loss}
\end{equation}
where $\mathcal{L}_{\mathrm{sem}}$ preserves parent--child semantic coherence, $\mathcal{L}_{\mathrm{temp}}$ encourages temporal and citation-direction consistency, and $\mathcal{L}_{\mathrm{marg}}$ trains the marginal detector (\S\ref{sec:3.4.3}). Detailed formulations in Appendix~\ref{app:posttrain_loss}.

\subsubsection{Structural Admissibility Constraints}
\label{sec:3.4.2}

In addition to the differentiable objectives, the E-step filters candidate trees using admissibility constraints. These constraints prevent the evolutionary refinement from producing structures that are temporally invalid or that destroy the semantic backbone learned in Stage I.

\paragraph{Temporal consistency.}
This constraint follows Eq.~\ref{eq:temporal-consistency}, directly rejecting violating merges.

\paragraph{Backbone preservation.}
This retains the major branches of $\mathcal{T}^{\mathrm{tax}}$ via the Major Branch Count:
\begin{equation}
\mathrm{MBC}(\mathcal{T}) = \big|\{c\in\mathrm{children}(r):|\mathcal{V}(c)|\geq\rho|\mathcal{V}|\}\big|,
\label{eq:mbc}
\end{equation}
requiring $\mathrm{MBC}(\mathcal{T}) \geq \kappa\,\mathrm{MBC}(\mathcal{T}^{\mathrm{tax}})$.

\paragraph{Tree legality.}
The output must remain a valid rooted tree, with a single root, no cycles, and a unique parent for each non-root concept node. Detailed constraints are described in Appendix~\ref{app:constraints}.

\subsubsection{Marginal Paper Handling}
\label{sec:3.4.3}

Stage~II explicitly handles \emph{marginal papers}: papers poorly explained by all fine-grained leaf concepts, which would blur concept boundaries and weaken temporal coherence if forced into leaves. We separate handling into leaf-level detection and all-node re-attachment.
 
\paragraph{Detection.}
For each paper $v_i$, we compute its best leaf compatibility $A_i^{\mathrm{leaf}} = \max_{c \in \mathrm{Leaves}(\mathcal{T}^{(t)})} m(v_i,c)$, where $m(v_i,c)\in[0,1]$ combines semantic similarity, citation-neighbor overlap, and temporal compatibility (Appendix~\ref{app:marginal}). A paper is marginal when $A_i^{\mathrm{leaf}}<\eta$, providing pseudo-labels for a marginal classifier (optionally reinforced by HDBSCAN low-membership outliers). Intuitively, a paper is marginal when it is not well explained by any single leaf: its compatibility with even its best-matching concept falls below $\eta$, typically because its citations span multiple sibling branches rather than concentrating in one.
 
\paragraph{Re-attachment.}
Each flagged marginal paper is re-attached to the concept whose Gaussian best explains its embedding:
\begin{equation}
a(v_m) = \arg\!\max_{c \in \mathcal{C}^{(t)}} \log \mathcal{N}\!\left(\mathbf{e}_m;\, \boldsymbol{\mu}_c, \mathrm{diag}(\boldsymbol{\sigma}_c^2)\right).
\label{eq:marginal-attach}
\end{equation}
Unlike detection, re-attachment searches over all concept nodes, allowing papers that fit no fine-grained leaf to attach to broader internal concepts.

\subsection{Stage III: Few-shot Metric Learning}
\label{sec:3.5}
To inject structural priors that self-supervised signals alone cannot recover, we calibrate the model on a small set of reference evolutionary trees via few-shot metric learning. Only a lightweight adaptation layer ($\sim$17\% of parameters) is updated under a small learning rate (5e-6) for 20 epochs over the reference trees (10 per leave-one-domain-out fold) --- orders of magnitude smaller than standard supervised fine-tuning. The calibration losses are purely metric-learning: they shape embedding geometry to reflect concept clusters and evolution directions, without predicting discrete labels.
\begin{equation}
\mathcal{L}^{\mathrm{cal}} = \mathcal{L}^{\mathrm{cal}}_{\mathrm{marg}} + \mathcal{L}^{\mathrm{cal}}_{\mathrm{concept}} + \mathcal{L}^{\mathrm{cal}}_{\mathrm{edge}}.
\label{eq:cal-loss}
\end{equation}
Weight and learning-rate choices ensure the reference signal refines rather than overrides earlier structure. Full formulations in Appendix~\ref{app:gt_loss}.

\subsection{Concept Labeling}\label{sec:3.6}
The tree $\mathcal{T}$ encodes structure but no human-readable labels. A single LLM pass traverses $\mathcal{T}$ bottom-up:
\begin{equation}
\small
\phi(c) = \mathrm{LLM}\big(\Pi(c,\, \mathcal{V}(c),\, \{\phi(c') : c' \in \mathrm{children}(c)\})\big),
\label{eq:concept-label}
\end{equation}
where $\Pi$ is a structured prompt. Leaf prompts request a precise method name; internal-node prompts require a more abstract noun phrase that generalizes over (and does not repeat) child labels. \textbf{The LLM is used only for concept verbalization and does not modify topology or paper attachments.} Prompt templates in Appendix~\ref{app:concept_prompts}.
\section{Experiments}
\label{sec:exp}

\subsection{Datasets}
\label{sec:datasets}

% We evaluate EvoTree on two datasets. The first is a large-scale survey-reference graph dataset used for representation pre-training, validation, and self-supervised evaluation. The second is a smaller annotated dataset used only for few-shot evaluation.

\paragraph{Survey-reference graph dataset.}
We construct 411 ego-graphs from arXiv survey papers. Hierarchical weak labels are derived from the survey section structure: a referenced paper inherits the section path in which it appears, e.g., \textit{Chapter 2} $\rightarrow$ \textit{2.2} $\rightarrow$ \textit{2.2.3}. We split graphs, rather than individual nodes, into train/validation/test sets with a 70/15/15 ratio. Detailed statistics are shown in Table~\ref{tab:dataset_main}.

% \begin{table}[t]
% \centering
% \small
% \caption{Statistics of the survey-reference graph dataset.}
% \label{tab:dataset_main}
% \begin{tabular}{l|r}
% \toprule
% \textbf{Statistic} & \textbf{Value} \\
% \midrule
% Graphs & 411 \\
% Paper nodes & 44,812 \\
% Citation edges & 205,725 \\
% Avg. nodes / graph & 109.0 \\
% Avg. edges / graph & 500.5 \\
% Node embedding dim. & 768 \\
% \bottomrule
% \end{tabular}
% \end{table}

\paragraph{Few-shot (FS) calibration dataset.}
For direct evaluation against annotated labels, we curate 11 reference evolutionary trees spanning major AI subfields, including computer vision, natural language processing, graph learning, reinforcement learning, and large language models. Each graph contains manual annotations for concept membership (\texttt{FS\_concept}), marginal-paper labels (\texttt{FS\_is\_marginal}), evolution edges (\texttt{FS\_tree\_edges}), and node depth (\texttt{FS\_depth}). Detailed statistics are shown in Table~\ref{tab:dataset_gt}. Calibration follows an 11-fold leave-one-domain-out protocol (Appendix~\ref{app:fs-overlap}).
% \begin{table}[t]
% \centering
% \small
% \caption{Statistics of the FS dataset.}
% \label{tab:dataset_gt}
% \begin{tabular}{l|r}
% \toprule
% \textbf{Statistic} & \textbf{Value} \\
% \midrule
% Domains & 11 \\
% Paper nodes & 352 \\
% FS evolution edges & 90 \\
% FS marginal papers & 24 \\
% Avg. nodes / domain & 32.0 \\
% \bottomrule
% \end{tabular}
% \end{table}

\subsection{Clustering Semantic Quality and Evolution Structure Quality}
\label{sec:exp_clustering_evolution}

\subsubsection{Metrics \& Baselines}

We evaluate two aspects of tree quality. For semantic clustering, we report \textbf{Leaf Purity (LP)} and \textbf{Path Purity (PP)} to measure topic consistency at the leaf and root-to-leaf path levels, respectively, and \textbf{NMI} to measure agreement between the tree-induced partition and topic labels. For evolution structure, we report \textbf{Citation Direction Accuracy (CDA)}, which checks whether cited papers are placed at shallower depths than citing papers, and \textbf{Path Monotonicity Rate (PMR)}, which measures whether publication years are non-decreasing along root-to-leaf paths.

We compare EvoTree with representative flat, hierarchical, citation-aware, and taxonomy-induction baselines, including HAC-SPECTER2~\cite{Murtagh2012AlgorithmsFH}, KMeans-flat~\cite{electronics9081295}, HAC-4level~\cite{Murtagh2012AlgorithmsFH}, CitRank+HAC~\cite{Woods2024APA}, TaxoGen-style~\cite{zhang2018taxogen}, Hu-CiteTaxo~\cite{hu2024hi}, TaxoAlign~\cite{lahiri2025taxoalign}, and Context-Aware~\cite{zhu-etal-2025-context}. We also report \textbf{EvoTree (Pre-train)}, which disables temporal fine-tuning, to isolate the effect of representation--structure refinement.

% -----------------------------------------------------------------------------
\subsubsection{Results}
% -----------------------------------------------------------------------------

% Table~\ref{tab:exp1_clustering_evolution} reports performance across all methods.

\begin{table*}[t]
\centering
\small
\caption{Clustering semantic quality and evolution structure quality. Dashes indicate that the method does not produce the required output. Results are mean $\pm$ standard deviation over 5 runs. Best results are in \textbf{bold}.}
\label{tab:exp1_clustering_evolution}
\begin{tabular}{l|ccc|cc}
\toprule
\multirow{2}{*}{\textbf{Method}} & \multicolumn{3}{c|}{\textbf{Clustering}} & \multicolumn{2}{c}{\textbf{Evolution}} \\
                                  & LP $\uparrow$ & PP $\uparrow$ & NMI $\uparrow$ & CDA $\uparrow$ & PMR $\uparrow$ \\
\midrule
HAC-SPECTER2                      & 0.526$\pm$0.008 & 0.607$\pm$0.009 & 0.399$\pm$0.011 & --              & --              \\
KMeans-flat                       & 0.524$\pm$0.010 & 0.603$\pm$0.011 & 0.392$\pm$0.012 & --              & --              \\
HAC-4level                        & 0.586$\pm$0.007 & 0.665$\pm$0.008 & 0.437$\pm$0.010 & 1.000  & 0.293$\pm$0.014 \\
CitRank+HAC                       & 0.498$\pm$0.011 & 0.574$\pm$0.012 & 0.346$\pm$0.013 & 1.000     & 0.250$\pm$0.015 \\
TaxoGen-style                     & 0.601$\pm$0.009 & 0.674$\pm$0.008 & 0.511$\pm$0.010 & 0.892$\pm$0.013 & 0.265$\pm$0.013 \\
\midrule
Hu-CiteTaxo 
& 0.664$\pm$0.006 
& 0.718$\pm$0.008 
& 0.519$\pm$0.005 
& -- 
& \textbf{0.322$\pm$0.007} \\

TaxoAlign 
& 0.650$\pm$0.007 
& 0.694$\pm$0.006 
& 0.480$\pm$0.009 
& 0.883$\pm$0.005 
& 0.305$\pm$0.008 \\

Context-Aware 
& \textbf{0.713$\pm$0.005} 
& \textbf{0.742$\pm$0.007} 
& 0.512$\pm$0.006 
& 0.821$\pm$0.009 
& 0.318$\pm$0.004 \\
\midrule
EvoTree (Pre-train)               & 0.591$\pm$0.008 & 0.689$\pm$0.009 & 0.494$\pm$0.011 & 0.829$\pm$0.012 & 0.284$\pm$0.013 \\
\textbf{EvoTree}                  & 0.653$\pm$0.007 & 0.703$\pm$0.008 & \textbf{0.526$\pm$0.009} & \textbf{0.893$\pm$0.011} & 0.312$\pm$0.012 \\
\bottomrule
\end{tabular}

\vspace{0.5em}
\raggedright
% \footnotesize
% $^{\dag}$ HAC-4level achieves CDA $=1.000$ as an artifact: the method does not consume citation information, so its CDA degenerates to a trivially passing case.
% $^{*}$ CitRank+HAC produces the same artifact for analogous reasons.
\end{table*}

\paragraph{Semantic clustering quality.}
Taxonomy-generation methods remain competitive here: Context-Aware leads on LP (0.713) and PP (0.742), exceeding EvoTree by 6.0 and 3.9 points. EvoTree attains the highest NMI (0.526). These metrics reward flat leaf coherence alone; EvoTree additionally constrains its hierarchy to be temporally admissible, trading some semantic compactness for evolution-consistent structure.

\paragraph{Citation- and time-consistent evolution structure.}
Hu-CiteTaxo attains the highest PMR (0.322); its CDA is not
comparable, as its output does not expose the directional edges the metric evaluates. The more informative distinction is structural: none of the
three taxonomy-generation baselines emits evolution edges or marginal scores, so they are not evaluated on the FS-aligned metrics (\S\ref{sec:exp_gt}). The indirect metrics above cover only the semantic objective that
taxonomy generation shares with evolution-tree induction.

\paragraph{Effect of temporal fine-tuning.}
Comparing EvoTree with EvoTree (Pre-train) isolates the effect of the temporal fine-tuning stage. Full EvoTree improves LP and CDA, showing that joint representation--structure refinement improves both semantic coherence and citation-direction consistency.
% =============================================================================
\subsection{Evaluation on Few-shot Test}
\label{sec:exp_gt}

% While \S\ref{sec:exp_clustering_evolution} evaluates tree quality using indirect signals (topic labels and citation edges), this section evaluates against labels extracted from high-impact survey papers.

\subsubsection{Metrics \& Baselines}

We evaluate against the annotated few-shot dataset using three metrics: \textbf{FS-CP}, which measures concept purity within predicted leaf nodes; \textbf{FS-EDA}, which measures whether annotated evolutionary descendants are placed deeper than their predecessors; and \textbf{FS-Mar}, which evaluates marginal-paper detection by AUROC.

We compare EvoTree with clustering baselines from \S\ref{sec:exp_clustering_evolution} and LLM-based baselines that prompt GPT-4o to generate hierarchies from titles, titles plus years, or titles plus citation-graph context. We again include EvoTree (Pre-train) to measure the effect of temporal fine-tuning.

\subsubsection{Results}
% -----------------------------------------------------------------------------

% Table~\ref{tab:exp2_gt} reports performance against the annotated dataset.

% EvoTree achieves the best score on FS-CP and the only non-trivial score on FS-Mar, while remaining competitive with the strongest LLM baseline on FS-EDA.

\begin{table*}[h!]
\centering
\small
\caption{Evaluation on the few-shot benchmark. Results are mean $\pm$ standard deviation over 5 runs. Best results are in \textbf{bold}.}
\label{tab:exp2_gt}
\begin{tabular}{l|ccc}
\toprule
\textbf{Method}            & FS-CP $\uparrow$       & FS-EDA $\uparrow$         & FS-Mar $\uparrow$       \\
\midrule
HAC-SPECTER2               & 0.600$\pm$0.011        & 0.838$\pm$0.013           & --                       \\
CitRank+HAC                & 0.739$\pm$0.010        & 1.000           & --                       \\
TaxoGen-style              & 0.661$\pm$0.011        & 0.929$\pm$0.012           & --                       \\
\midrule
LLM (title)                & 0.767$\pm$0.009        & 0.964$\pm$0.010           & --                       \\
LLM (title + year)         & 0.759$\pm$0.010        & 0.937$\pm$0.011           & --                       \\
LLM (citation graphs)      & 0.769$\pm$0.009        & 0.952$\pm$0.010           & --                       \\
\midrule
EvoTree (Pre-train)        & 0.653$\pm$0.011        & 0.904$\pm$0.012           & 0.662$\pm$0.013          \\
\textbf{EvoTree}           & \textbf{0.775$\pm$0.008} & 0.949$\pm$0.010         & \textbf{0.726$\pm$0.011} \\
\bottomrule
\end{tabular}

\vspace{0.5em}
\raggedright
% \footnotesize
% $^{\dag}$ CitRank+HAC achieves FS-EDA $=1.000$ as an artifact analogous to that observed in Table~\ref{tab:exp1_clustering_evolution}: the method's evolution-direction check degenerates to a trivially passing case under its citation-rank-based scoring.
\end{table*}

\paragraph{Alignment with annotated concepts.}
On the few-shot dataset, EvoTree achieves the highest FS-CP, though the gap is narrow. Its improvement over the strongest LLM baseline is modest, but it consistently outperforms clustering-based baselines and substantially improves over EvoTree (Pre-train), indicating that temporal fine-tuning and tree refinement help align predicted clusters with ground-truth concepts.

\paragraph{Evolution-direction accuracy.}
EvoTree is on par with the strongest LLM baseline on FS-EDA. The strong performance of LLM-title suggests that large language models can often infer local predecessor--successor relations from method names alone. However, LLM-only methods do not provide explicit structure-level constraints or marginality estimates, which are required for controlled, audit-friendly tree construction.

\paragraph{Marginal-paper detection.}
FS-Mar reflects a structural capability that, to our knowledge, no other method evaluated in this work exposes:
standard clustering and LLM baselines produce only hard assignments and therefore admit no continuous marginality score, while EvoTree directly outputs marginality scores from its hierarchical structure.
EvoTree obtains a non-trivial FS-Mar AUROC and improves over its pre-training-only variant, supporting the value of modeling boundary papers rather than forcing every paper into a fine-grained leaf cluster.

%\paragraph{Summary.}
%Taken together, the two experiments show that EvoTree (i) produces tighter, semantically more coherent clusters than embedding-based and term-based hierarchical clustering methods; (ii) recovers the directional evolution structure of methodological progress as evidenced by both citation-based and expert-annotated signals; and (iii) is the only method evaluated that exposes a meaningful marginal-paper score, enabling boundary-aware taxonomy curation. The remaining experiments isolate the contribution of each design component (\S\ref{sec:ablation}) and study sensitivity to key hyperparameters (\S\ref{sec:param_study}).

\subsection{Ablation Studies}
\label{sec:ablation}

\subsubsection{Contribution of Each Training Stage}
\label{sec:ablation_training}

EvoTree is trained in three stages: \textbf{Pre-train only} (Stage~I) uses multi-task supervision with section structure as weak labels; \textbf{+\,Post-train} (Stage~II) adds self-supervised evolutionary losses; \textbf{+\,FS-correction} (Stage~III few-shot calibration; our full model) injects structural priors via few-shot metric learning on the reference trees. We evaluate on test graphs and on the FS dataset (Table~\ref{tab:abl_training}) to verify that gains generalize beyond the FS-correction training set.

\begin{table*}[t]
\centering
\small
\setlength{\tabcolsep}{3pt}
\caption{Ablation study on the contribution of each training stage. Left: clustering and evolution-structure metrics on test graphs; right: FS-aligned metrics on the 11 annotated domains. Results are mean $\pm$ standard deviation over 5 runs. Best results are in \textbf{bold}.}
\label{tab:abl_training}
\begin{tabular}{l|ccc|cc|ccc}
\toprule
\multirow{2}{*}{\textbf{Variant}} & \multicolumn{3}{c|}{\textbf{Clustering}} & \multicolumn{2}{c|}{\textbf{Evolution}} & \multicolumn{3}{c}{\textbf{FS-aligned}} \\
 & LP $\uparrow$ & PP $\uparrow$ & NMI $\uparrow$ & CDA $\uparrow$ & PMR $\uparrow$ & FS-CP $\uparrow$ & FS-EDA $\uparrow$ & FS-Mar $\uparrow$ \\
\midrule
Pre-train only       & 0.541\std{0.010} & 0.629\std{0.011} & \textbf{0.530\std{0.011}} & 0.829\std{0.013} & 0.240\std{0.014} & 0.553\std{0.011} & \textbf{0.954\std{0.012}} & 0.662\std{0.013} \\
\;\;+\,Post-train    & 0.598\std{0.009} & 0.665\std{0.009} & 0.455\std{0.011} & 0.857\std{0.012} & 0.270\std{0.013} & 0.657\std{0.010} & 0.880\std{0.012} & \textbf{0.750\std{0.012}} \\
\;\;+\,FS-correction & \textbf{0.653\std{0.007}} & \textbf{0.703\std{0.008}} & 0.526\std{0.009} & \textbf{0.893\std{0.011}} & \textbf{0.312\std{0.012}} & \textbf{0.775\std{0.008}} & 0.949\std{0.010} & 0.726\std{0.011} \\
\bottomrule
\end{tabular}
\end{table*}

Both later stages deliver substantial gains in leaf purity and concept-level alignment (FS-CP $+0.104$ after post-training and a further $+0.118$ after FS-correction), consistent with the self-supervised evolutionary losses injecting temporal structure that label supervision cannot capture, and with the few-shot priors sharpening concept boundaries. Post-training yields the largest gain in marginal-paper recognition (FS-Mar $0.662 \rightarrow 0.750$), most of which FS-correction preserves ($0.726$). These gains also transfer to test graphs unseen during FS-correction training, mitigating concerns about overfitting to the reference trees. The dip in FS-EDA after post-training reflects the tension between self-supervised temporal refinement and strict depth ordering; FS-correction recovers most of it ($0.880 \rightarrow 0.949$).

\subsubsection{Necessity of Graph Structure}
\label{sec:ablation_graph}

To verify that the GNN exploits the citation graph rather than merely propagating SPECTER2 embeddings, we construct \textbf{Pre-train (MLP)} --- identical to \textbf{Pre-train (GNN)} in hyperparameters and training objective but with the citation edge set replaced by an empty set, so any gap is attributable solely to the citation graph.

\begin{table*}[t]
\centering
\small
\caption{Ablation study on the necessity of the citation graph. ``Val.\ score'' is the validation score during training. Best in \textbf{bold}. Mean $\pm$ std over 5 runs.}
\label{tab:abl_graph}
\begin{tabular}{l|cccc|c}
\toprule
\textbf{Variant}             & LP $\uparrow$ & PP $\uparrow$ & NMI $\uparrow$ & PMR $\uparrow$ & Val.\ score $\uparrow$ \\
\midrule
Pre-train (MLP, no graph)    & 0.450$\pm$0.012 & 0.536$\pm$0.013 & 0.363$\pm$0.013 & \textbf{0.306$\pm$0.013} & 0.514$\pm$0.014 \\
Pre-train (GNN, with graph)  & \textbf{0.541$\pm$0.010} & \textbf{0.629$\pm$0.011} & \textbf{0.464$\pm$0.011} & 0.240$\pm$0.014 & \textbf{0.780$\pm$0.010} \\
\bottomrule
\end{tabular}
\end{table*}

\begin{table*}[h!]
\centering
\small
\caption{Generation-quality evaluation on the FS domains under two independent protocols, LLM-as-a-Judge and Human evaluation. Scores are on a 1--10 scale (mean $\pm$ standard deviation over LLM runs or human evaluators). Best results are in \textbf{bold}.}
\setlength{\tabcolsep}{4pt}
\begin{tabular}{lcccc|cccc}
\toprule
& \multicolumn{4}{c|}{LLM-as-a-Judge} & \multicolumn{4}{c}{Human evaluation} \\
\cmidrule(lr){2-5}\cmidrule(lr){6-9}
Method & Concept & Evolution & Transitional & Overall & Concept & Evolution & Transitional & Overall \\
\midrule
CitRank+HAC   & 4.21$\pm$0.12 & 3.57$\pm$0.14 & 2.82$\pm$0.06 & 4.08$\pm$0.04 & 4.77$\pm$0.07 & 4.38$\pm$0.05 & 4.90$\pm$0.09 & 4.21$\pm$0.09 \\
Hu-CiteTaxo   & 4.55$\pm$0.11 & 4.73$\pm$0.07 & 3.57$\pm$0.13 & 4.64$\pm$0.14 & 6.65$\pm$0.15 & 5.87$\pm$0.10 & 5.82$\pm$0.01 & 6.19$\pm$0.05 \\
TaxoAlign     & 5.44$\pm$0.05 & 4.89$\pm$0.11 & 4.22$\pm$0.05 & 4.67$\pm$0.06 & 5.94$\pm$0.05 & 5.61$\pm$0.08 & 5.26$\pm$0.05 & 6.44$\pm$0.04 \\
Context-Aware & 5.60$\pm$0.14 & 5.20$\pm$0.04 & 4.40$\pm$0.07 & 5.10$\pm$0.12 & 6.81$\pm$0.04 & 6.32$\pm$0.14 & 5.97$\pm$0.01 & 6.67$\pm$0.09 \\
TaxoGen-style & \textbf{7.14$\pm$0.05} & 5.66$\pm$0.11 & 4.97$\pm$0.04 & 6.27$\pm$0.12 & \textbf{8.64$\pm$0.05} & 6.18$\pm$0.03 & 5.93$\pm$0.13 & 7.40$\pm$0.12 \\
EvoTree       & 6.03$\pm$0.13 & \textbf{7.51$\pm$0.04} & \textbf{6.85$\pm$0.10} & \textbf{6.82$\pm$0.05} & 7.16$\pm$0.08 & \textbf{8.65$\pm$0.15} & \textbf{6.47$\pm$0.11} & \textbf{8.05$\pm$0.03} \\
\bottomrule
\end{tabular}
\label{tab:genqual}
\end{table*}

Removing the citation graph causes an even larger gap in validation score than at inference, indicating that the graph contributes signal throughout optimization. The MLP variant's slightly higher PMR is a degenerate artifact: without citation constraints, tree construction relies on the temporal distribution of embeddings alone, trivially producing monotonic paths --- mirroring the CDA artifact for HAC-4level in \S\ref{sec:exp_clustering_evolution}. Together with the training-stage ablation (\S\ref{sec:ablation_training}), these results confirm that EvoTree relies jointly on the citation graph as a structural backbone and on the staged training framework that progressively injects supervised, self-supervised, and correction signals.

\subsection{Generation Quality}
The metrics in Tables~\ref{tab:exp1_clustering_evolution} and~\ref{tab:exp2_gt} measure agreement with reference partitions and citation ordering, but not whether a structure reads as a coherent account of how a field developed. We therefore evaluate the generated structures directly under two independent protocols: an LLM judge, instantiated with the same model as concept labeling (\S\ref{sec:3.6}) and run multiple times per structure, and three CS graduate students distinct from the FS annotators, scoring independently. Both apply the same four criteria---conceptual organization, scientific evolution, transitional-paper placement, and overall quality---to anonymized structures in randomized order. Protocol details are in Appendix~\ref{app:genqual}.

Both protocols rank EvoTree first on overall quality and produce the same overall ordering of all six methods (Table~\ref{tab:genqual}). EvoTree leads on scientific evolution and transitional-paper placement, the two dimensions taxonomy induction does not target, while TaxoGen-style leads on conceptual organization. The latter follows from temporal admissibility: when a method family develops over a long span, enforcing Eq.~\ref{eq:temporal-consistency} separates papers that are semantically adjacent but temporally distant, at some cost to leaf-level compactness---the same trade-off behind EvoTree's lower LP and PP. The rubric is generic across hierarchy-generation methods and does not reward EvoTree-specific mechanisms such as temporal constraints or internal-node re-attachment.

\section{Conclusion}
We introduced EvoTree, the first framework for automatically inducing scientific evolution trees from citation graphs, and formalized the task with an annotated benchmark covering 11 AI subfields. By decoupling conceptual backbone learning from temporal refinement, EvoTree first constructs a stable taxonomy via graph-aware encoding and distribution-based hierarchical clustering, then re-attaches marginal papers to internal nodes under monotonic-path constraints, and finally invokes an LLM solely for concept verbalization without altering topology. Against taxonomy-induction, citation-only, and LLM-based baselines, EvoTree attains the best NMI, citation-direction accuracy, and annotated concept purity, is preferred by both LLM and human judges on overall quality, and is the only method that recovers transitional papers with non-trivial accuracy; ablations confirm that both the citation graph and the staged training framework are jointly necessary. We view evolution-tree induction as a step toward scalable, temporally faithful organization of scientific literature.

\section*{Limitations}

\paragraph{Scope of domains.}
Our 411 ego-graphs and 11 reference trees are drawn from AI surveys. We have not verified transfer to disciplines whose citation conventions and paradigm cadences differ substantially from AI.

\paragraph{Scale of the few-shot reference set.}
The FS dataset comprises 11 curated reference trees and 352 papers, sufficient to demonstrate the few-shot regime but limiting fine-grained per-domain analyses; a larger curated set would strengthen alignment evaluation.

\paragraph{Reliance on survey section structure.}
Stage~I assumes that section organization reflects a coherent conceptual taxonomy. Surveys organized by application area or chronology may provide weaker supervision than those organized by methodology.

\paragraph{Temporal signal noise.}
Publication years conflate arXiv preprint, conference, and journal dates, introducing noise into the temporal-admissibility constraint and PMR. This contributes to the modest absolute PMR values observed in our experiments.

\paragraph{External LLM dependence.}
Concept labels are generated by GPT-4.1-mini. Although the LLM is used only for verbalization and does not alter topology, this introduces a dependency on a closed-source model; substituting an open-source LLM is straightforward but may change labeling style.

\bibliography{custom}

\clearpage
\appendix

% ============================================================
% Required packages (add to preamble if not already present):
% \usepackage{longtable}
% \usepackage{booktabs}
% \usepackage{array}
% \usepackage{hyperref}
% ============================================================

\section{Case Study: Evolution Tree on Adversarial Attacks on Text-to-Image Diffusion Models}
\label{app:case-study}

\subsection{Setup}
\label{app:case-study:setup}

We instantiate EvoTree on the reference corpus of \emph{Adversarial Attacks and Defenses on Text-to-Image Diffusion Models: A Survey}~\cite{zhang2025adversarial}\footnote{arXiv:2407.15861.}, which serves as both the ego node and the topical anchor of the resulting tree. The survey was selected because (i) it provides a recent, curated, and topically coherent snapshot of an actively evolving subfield, (ii) it spans both attack-side and defense-side work, allowing the tree to surface convergent research lines, and (iii) the ego node itself appears as paper [20] in the corpus, providing a natural reference point for the green branch. Unlike the experiments setting, we retain the survey node in this case-study visualization to anchor the visual layout and to illustrate how the ego work would be positioned by EvoTree in its own subfield.

\begin{figure*}[h]
  \centering
  \includegraphics[width=1\linewidth]{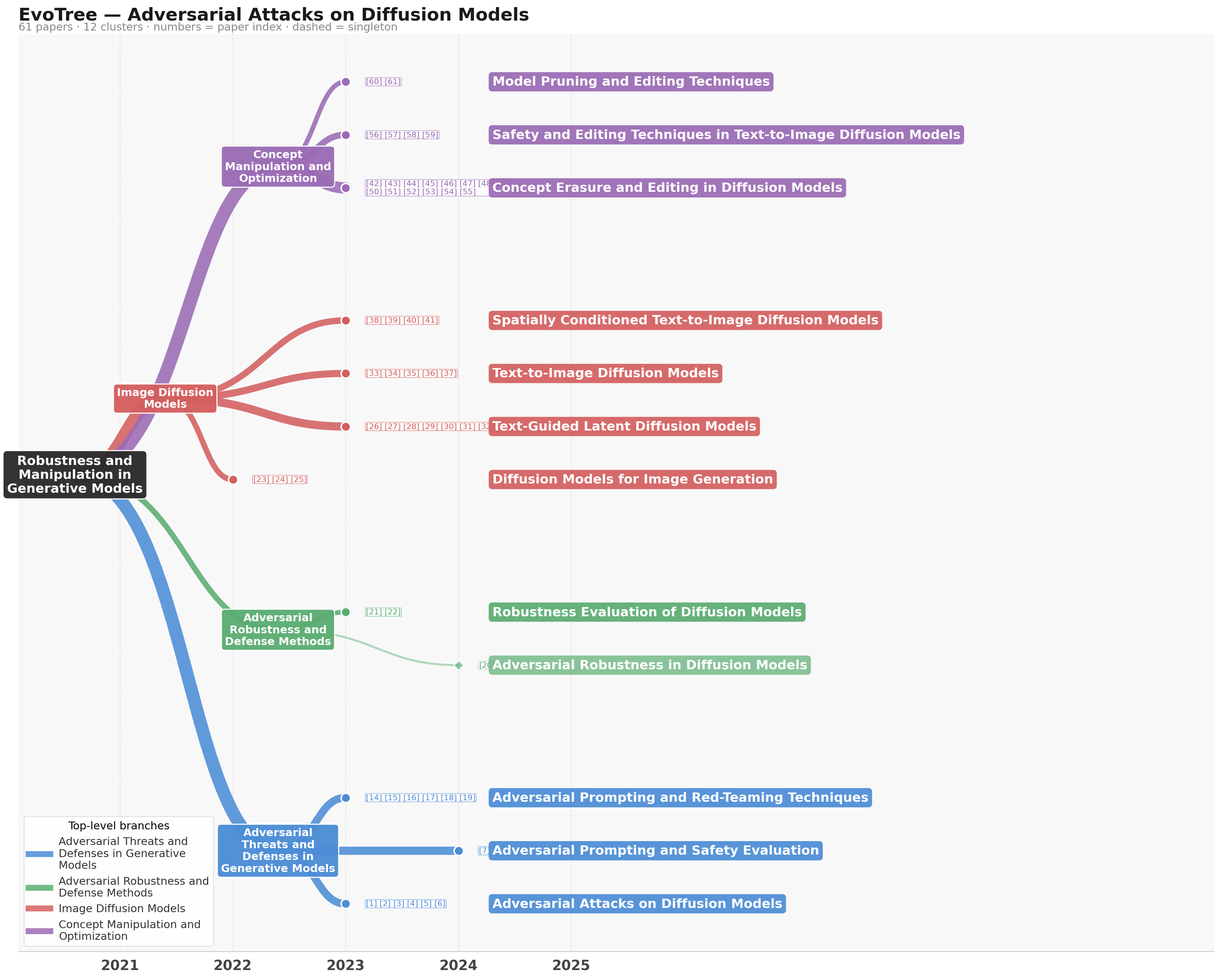}
  \caption{Evolution tree generated by EvoTree for the topic
\emph{Adversarial Attacks on Text-to-Image Diffusion Models}, instantiated on the reference corpus of the survey by \citet{zhang2025adversarial}. Cluster labels are produced by an LLM; the layout is rendered from the induced tree.}
  \label{fig:evotree-case}
\end{figure*}

After deduplication and minor filtering of off-topic citations, the corpus contains $61$ papers published between 2021 and 2024. EvoTree organizes them into $12$ leaf clusters grouped under $4$ top-level branches, as visualized in Figure~\ref{fig:evotree-case}. The bracketed indices $[1]$--$[61]$ refer exclusively to the corpus entries and are independent of the bibliography numbering in the main paper.

\subsection{Branch and Cluster Overview}
\label{app:case-study:branches}

The four top-level branches partition the corpus along complementary dimensions of the research landscape:
\begin{itemize}
    \item \textbf{Adversarial Threats and Defenses in Generative Models} (19 papers, blue) --- work targeting T2I diffusion models with adversarial prompts, red-teaming protocols, and safety filter evaluation.
    \item \textbf{Adversarial Robustness and Defense Methods} (3 papers, green) --- robustness evaluation studies and the ego survey itself.
    \item \textbf{Text-Conditioned Image Diffusion Models} (19 papers, red) --- the foundation-model lineage on which downstream attack/defense work depends, from unconditional DDPM-class models through latent and spatially conditioned T2I systems.
    \item \textbf{Concept Manipulation and Optimization} (20 papers, purple) --- concept erasure, model editing, and pruning techniques that intersect heavily with the safety agenda.
\end{itemize}

\subsection{Analysis and Discussion}
\label{app:case-study:analysis}

EvoTree recovers three structural properties of the subfield. (i) A clear \emph{foundation--application} stratification: the red branch (DiT~[24] \cite{peebles2023scalable}, latent diffusion~[31]\cite{rombach2022high}, GLIDE~[32]\cite{nichol2021glide}, SDXL~[27]\cite{podell2024sdxl}, and spatially conditioned variants~[38]--[41]\cite{couairon2023zero, chefer2023attend, avrahami2023spatext, balaji2022ediff}) temporally and conceptually precedes the attack and editing work in the blue and purple branches. (ii) The concept-erasure cluster (entries [42]--[55]\cite{wu2025unlearning, zhang2024defensive}) is the densest single thematic block, marking it as the dominant defense paradigm in this timeframe---ahead of input-side filtering or output-side detection. (iii) The blue branch is essentially confined to 2023 and later, placing adversarial-attack research on T2I~\cite{yang2024guardt2i} roughly two years behind the foundation work.

% \paragraph{Limitations.}
% Two structural caveats apply. (i) \emph{Convergent dependencies are not natively expressible in a tree.} The blue, green, and purple branches all presuppose the red one, yet diverge in parallel from a single root; temporal positioning of cluster centroids (red $\approx$ 2021, blue $\approx$ 2023) only partially compensates. (ii) \emph{Top-level branches mix taxonomic axes:} the red branch is organized by \textbf{model type} (the object of study), while blue/green/purple are organized by \textbf{operation type} (what one does to that object)---a byproduct of HDBSCAN clustering on content embeddings without an enforced relational schema.

% \paragraph{Interpretive guidance.}
% EvoTree's output is best read as a \emph{facet map} of a subfield, with the foundation branch serving as an implicit trunk and operation-oriented branches as derivative offshoots. Parallel top-level branches should be treated as potentially convergent rather than independent, especially when the corpus mixes foundation work with derivative attack/defense work.

\section{Details of Datasets}
\label{app:gt_dataset}

\subsection{Main Dataset}
We construct 411 ego-graphs from arXiv survey papers. Each graph is centered on one survey paper: its references form the graph nodes, and citations among these references form the directed edges. The survey node itself is excluded. Each paper node carries a title, an abstract, a publication year, and a 768-dimensional SPECTER2~\cite{Singh2022SciRepEvalAM} embedding. Weak hierarchical labels are inherited from the survey's section structure: a referenced paper takes on the section path in which it appears, e.g., \textit{Chapter 2} $\rightarrow$ \textit{2.2} $\rightarrow$ \textit{2.2.3}. We split at the graph level, rather than the node level, into train/validation/test sets with a 70/15/15 ratio.

\begin{table}[h]
\centering
\small
\caption{Statistics of the survey-reference graph dataset.}
\label{tab:dataset_main}
\begin{tabular}{l|r}
\toprule
\textbf{Statistic} & \textbf{Value} \\
\midrule
Graphs & 411 \\
Paper nodes & 44,812 \\
Citation edges & 205,725 \\
Avg. nodes / graph & 109.0 \\
Avg. edges / graph & 500.5 \\
Node embedding dim. & 768 \\
\bottomrule
\end{tabular}
\end{table}

\subsection{Few-shot Dataset}
\begin{table}[h]
\centering
\small
\caption{Statistics of the FS dataset.}
\label{tab:dataset_gt}
\begin{tabular}{l|r}
\toprule
\textbf{Statistic} & \textbf{Value} \\
\midrule
Domains & 11 \\
Paper nodes & 352 \\
FS evolution edges & 90 \\
FS marginal papers & 24 \\
Avg. nodes / domain & 32.0 \\
\bottomrule
\end{tabular}
\end{table}

\paragraph{Annotation schema.}
Each domain graph contains four types of annotations:
(i) \texttt{FS\_concept}, which assigns each paper to a defined method concept;
(ii) \texttt{FS\_is\_marginal}, which marks papers that lie on topic boundaries or cannot be confidently assigned to a fine-grained concept;
(iii) \texttt{FS\_tree\_edges}, which records directed method-evolution relations, each typed as \texttt{extends}, \texttt{improves}, or (cross-domain) \texttt{adapts}, where the source paper is regarded as an extension or improvement of the target paper; and
(iv) \texttt{FS\_depth}, which indicates the depth of a paper in the evolution structure.

\paragraph{Domain coverage.}
The FS dataset contains 11 high-quality annotated AI subfields. We retain domains with sufficiently informative annotations for concept membership, evolution relations, and marginal-paper evaluation, and exclude domains whose evolution-edge annotations are too sparse for reliable FS-based comparison. Table~\ref{tab:app_gt_domains} reports the per-domain statistics.

\begin{table*}[h]
\centering
\small
\caption{Per-domain statistics of the annotated FS dataset. $N$ denotes the number of papers in each domain. ``Concepts'' denotes the number of annotated method concepts. ``FS edges'' denotes the number of annotated method-evolution edges. ``Marginal'' denotes the number of papers annotated as marginal or boundary papers.}
\label{tab:app_gt_domains}
\begin{tabular}{lrrrr}
\toprule
\textbf{Domain} & \textbf{$N$} & \textbf{Concepts} & \textbf{FS edges} & \textbf{Marginal} \\
\midrule
3D Vision & 25 & 5 & 11 & 4 \\
CV ConvNets & 40 & 7 & 4 & 1 \\
Vision Transformers & 27 & 4 & 2 & 3 \\
GAN Image Synthesis & 23 & 6 & 10 & 3 \\
LLM Alignment & 55 & 5 & 12 & 1 \\
Meta / Few-shot & 27 & 5 & 13 & 2 \\
NLP Transformers & 35 & 6 & 3 & 2 \\
Object Detection & 38 & 6 & 3 & 1 \\
Self-supervised Learning & 29 & 5 & 11 & 2 \\
Semantic Segmentation & 37 & 5 & 14 & 2 \\
Video Understanding & 16 & 6 & 7 & 3 \\
\midrule
\textbf{Total} & \textbf{352} & \textbf{60} & \textbf{90} & \textbf{24} \\
\textbf{Average} & \textbf{32.0} & \textbf{5.5} & \textbf{8.2} & \textbf{2.2} \\
\bottomrule
\end{tabular}
\end{table*}

\paragraph{FS evaluation protocol.}
The FS benchmark participates in Stage~III calibration under an 11-fold leave-one-domain-out protocol. In each fold, calibration uses the 10 remaining domains and evaluation is performed on the held-out domain; the held-out domain's concept labels, marginal-paper labels, evolution edges, and depth annotations are used neither for calibration nor for validation, threshold selection, or hyperparameter tuning. Each of the 11 domains serves once as the held-out domain; FS-aligned metrics are averaged over the 11 held-out evaluations, and we report the mean and standard deviation of this average over 5 runs.

\subsection{FS Cross-domain Overlap}
\label{app:fs-overlap}

The leave-one-domain-out protocol assumes that the held-out domain is genuinely unseen during calibration. Because the 11 FS domains are drawn from surveys within a single broad field, this assumption could be weakened if the same papers recur across domains: a paper appearing in both the calibration and held-out sets would leak its annotation indirectly. We therefore quantify paper-level overlap across all domain pairs.

Table~\ref{tab:overlap} reports the number of papers shared by each pair of domains, with domain sizes on the diagonal. Overlap is sparse throughout: the largest shared count is 3 papers, or 8.6\% when normalized by the smaller domain in the pair, and most pairs share at most one or two papers. % TODO(authors): verify. D-indices were inferred from the diagonal sizes vs. Table~\ref{tab:app_gt_domains}: D1 3D Vision, D2 CV ConvNets, D3 Vision Transformers, D4 GAN, D5 Meta/Few-shot, D6 Object Detection, D7 SSL, D8 Semantic Segmentation, D9 Video, D10 NLP Transformers, D11 LLM Alignment (D3/D5 both have 27 papers).
The non-negligible overlaps concentrate between semantically adjacent domains (D2/D6/D8: CV ConvNets, Object Detection, and Semantic Segmentation; D10/D11: NLP Transformers and LLM Alignment), which is expected since neighboring subfields cite a common set of foundational works. Under this level of overlap, a held-out domain's evaluation set is composed almost entirely of papers absent from the corresponding calibration folds, so the FS-aligned results are unlikely to reflect within-set memorization.

\begin{table}[h!]
\centering
\small
\setlength{\tabcolsep}{2pt}
\begin{tabular}{lccccccccccc}
\toprule
 & D1 & D2 & D3 & D4 & D5 & D6 & D7 & D8 & D9 & D10 & D11 \\
\midrule
D1  & \textbf{25} & 1 & 1 & 0 & 0 & 1 & 0 & 1 & 0 & 0 & 0 \\
D2  & 1 & \textbf{40} & 2 & 1 & 2 & 3 & 2 & 3 & 2 & 0 & 0 \\
D3  & 1 & 2 & \textbf{27} & 1 & 1 & 1 & 2 & 1 & 1 & 2 & 1 \\
D4  & 0 & 1 & 1 & \textbf{23} & 0 & 0 & 1 & 0 & 0 & 0 & 0 \\
D5  & 0 & 2 & 1 & 0 & \textbf{27} & 1 & 0 & 0 & 1 & 0 & 0 \\
D6  & 1 & 3 & 1 & 0 & 1 & \textbf{38} & 1 & 3 & 2 & 0 & 0 \\
D7  & 0 & 2 & 2 & 1 & 0 & 1 & \textbf{29} & 1 & 0 & 0 & 0 \\
D8  & 1 & 3 & 1 & 0 & 0 & 3 & 1 & \textbf{37} & 1 & 0 & 0 \\
D9  & 0 & 2 & 1 & 0 & 1 & 2 & 0 & 1 & \textbf{16} & 0 & 0 \\
D10 & 0 & 0 & 2 & 0 & 0 & 0 & 0 & 0 & 0 & \textbf{35} & 3 \\
D11 & 0 & 0 & 1 & 0 & 0 & 0 & 0 & 0 & 0 & 3 & \textbf{55} \\
\bottomrule
\end{tabular}
\caption{Pairwise paper-level overlap across the 11 FS domains. Diagonal entries (bold) give domain sizes; off-diagonal entries give the number of papers shared by the two domains.}
\label{tab:overlap}
\end{table}

We note that overlap counts alone do not exclude all forms of information sharing---semantically similar papers may appear under different identifiers across surveys. The reported figures bound direct duplication, not conceptual proximity between domains.

\subsection{FS Benchmark Construction and Annotation}
\label{app:fs-annotation}

The 11 reference evolution trees were curated from survey papers that explicitly present evolution diagrams or method-lineage organizations, rather than topic-only taxonomies. This requirement restricts the candidate pool but ensures each reference tree reflects an author-endorsed account of how the field developed, which we standardize into a unified annotation format.

Three graduate students annotated four aspects independently, following a shared guideline: concept membership (which concept node each paper belongs to), evolution edges (directed development relations between concepts), marginal-paper labels (whether a paper is better explained at a coarser abstraction level), and paper depth. Annotators worked from paper titles, abstracts, publication years, and the citation graph, and had access to the source survey.

\section{Concept-Labeling Prompt Templates}
\label{app:concept_prompts}
 
This appendix provides the full prompt templates used in the bottom-up concept generation pass described in \S\ref{sec:3.6}.
Both templates are instantiated per node and consumed by GPT-4.1-mini at temperature $0.1$.

\subsection{Leaf Prompt}
\label{app:leaf_prompt}
 
A leaf node represents a specific research micro-topic.
The prompt supplies all paper titles with publication years and instructs the model to produce a precise 3--6 word noun phrase naming the shared method or technique, explicitly discouraging generic openers (e.g., \emph{Advanced}, \emph{Large}).

\subsection{Internal-Node Prompt}
\label{app:internal_prompt}
 
An internal node must generalize over its children.
The prompt provides the already-assigned child concept names (bottom-up guarantee) and a sample of representative papers.
The model is explicitly required to produce a label that (i) is more abstract than any single child, and (ii) does not repeat any child label verbatim.

\begin{tcolorbox}
[colback=gray!5, colframe=gray!50, title=Leaf-node prompt, fonttitle=\small\bfseries, fontupper=\small\itshape, width=\columnwidth,]
You are assigning a precise concept label to a leaf cluster in an academic evolutionary taxonomy. The cluster groups closely related papers that share a specific research method, task, or technique.
 
\medskip
Papers in this cluster: \texttt{[list of titles with years]}
 
\medskip
Note: these papers span \texttt{YYYY--YYYY}. Reflect the temporal focus if it is distinctive.
 
\medskip
Requirements:
\begin{enumerate}
    \item Use 3--6 words; noun phrase, no verbs.
    \item Be specific --- prefer method/technique names over broad area names.
    \item Do NOT start with `Advanced', `Large', or `Modern'.
\end{enumerate}
Reply with ONLY the concept name, nothing else.
\end{tcolorbox}
 
\begin{tcolorbox}[colback=gray!5, colframe=gray!50, title=Internal-node prompt, fonttitle=\small\bfseries, fontupper=\small\itshape,width=\columnwidth]
You are assigning a concept label to an internal node in an academic evolutionary taxonomy. This node is the common parent of several sub-topics and must be named at a higher level of abstraction than its children.
 
\medskip
Child sub-topics: \texttt{[\textbullet\ child\_label\_1, \textbullet\ child\_label\_2, \ldots]}
 
\medskip
This subtree covers \texttt{YYYY--YYYY}.
 
\medskip
Representative papers from this subtree (sample): \texttt{[up to 5 titles with years]}
 
\medskip
Requirements:
\begin{enumerate}
    \item Use 3--6 words; noun phrase, no verbs.
    \item The name must generalize over ALL child sub-topics listed above.
    \item It must be more abstract than any single child concept.
    \item Do NOT repeat a child concept name verbatim.
\end{enumerate}
Reply with ONLY the concept name, nothing else.
\end{tcolorbox}

\section{Evaluation Metrics}\label{app:metrics}

This appendix specifies the eight evaluation metrics used in \S\ref{sec:exp}. We organize them into two parts: \emph{indirect metrics} computed on citation graphs without annotation, and \emph{few-shot metrics} computed on the 11 annotated graphs. Table~\ref{tab:metric-summary} summarizes the role of each metric.

\begin{table*}[h]
\centering
\small
\caption{Summary of evaluation metrics. ``FS'' indicates whether the metric requires annotation.}
\label{tab:metric-summary}
\begin{tabular}{l l l l c c}
\toprule
\textbf{Metric} & \textbf{What it measures} & \textbf{Signal source} & \textbf{Eq.} & \textbf{Stage} & \textbf{FS?} \\
\midrule
LP      & Leaf-cluster topic homogeneity        & Section labels       & Eq.~\ref{eq:lp}    & 1 & --- \\
PP      & Leaf-cluster path consistency         & Section paths        & Eq.~\ref{eq:pp}    & 1 & --- \\
NMI     & Cluster--class agreement at depth 2   & Section labels       & Eq.~\ref{eq:nmi}   & 1 & --- \\
CDA     & Citation direction vs.\ tree depth    & Citation graph       & Eq.~\ref{eq:cda}   & 2 & --- \\
PMR     & Whole-path temporal monotonicity      & Publication years    & Eq.~\ref{eq:pmr}   & 2 & --- \\
\midrule
FS-CP   & Leaf vs.\ concept              & Concepts      & Eq.~\ref{eq:FS-CP} & 3 & \checkmark \\
FS-EDA  & Evolution direction vs.\ tree depth   & Evolution edges & Eq.~\ref{eq:FS-EDA} & 3 & \checkmark \\
FS-Mar  & Marginal-paper detection (AUROC)      & Marginal flags & Eq.~\ref{eq:FS-Mar} & 3 & \checkmark \\
\bottomrule
\end{tabular}
\end{table*}

Throughout this appendix, $\mathcal{T}$ denotes a predicted evolutionary tree with leaf set $\mathrm{Leaves}(\mathcal{T}) = \{L_1, \ldots, L_K\}$. For each leaf $L_k$, $\mathcal{M}_k$ denotes the set of papers attached to $L_k$ (i.e., $\mathcal{V}(L_k)$ restricted to direct attachment). Each paper $v_i$ carries a publication year $\tau(v_i)$, a section-label set $\mathbf{y}_i \subseteq \mathcal{Y}$ (multi-hot weak labels from the survey vocabulary), and a section-path set $\mathbf{h}_i \subseteq \mathcal{H}$ from the pretrained section taxonomy. For papers in the annotated set $\mathcal{V}^{\mathrm{fs}}$, additional labels include a concept $g_i \in \mathcal{G}^{\mathrm{fs}}$ (single-label, deterministic), a marginal flag $y_i^{\mathrm{m,fs}} \in \{0,1\}$, and an evolution edge set $\mathcal{E}^{\mathrm{evo}}$ over papers. We write $\delta(v_i)$ for the depth of the concept node to which $v_i$ is attached --- whether that node is a leaf or an internal node (root depth $= 0$).
For methods that produce only leaf-level partitions (all baselines in this work), $\delta(v_i)$ is trivially the depth of the assigned leaf.

\subsection{Indirect metrics on held-out test graphs}

\paragraph{LP --- Leaf Label Purity.}
The label purity of a leaf $L_k$ is the fraction of its labeled papers covered by its single most frequent section label:
\begin{equation}
\mathrm{Purity}(L_k) = \frac{\displaystyle\max_{\ell \in \mathcal{Y}} \big|\{v_i \in \mathcal{M}_k : \ell \in \mathbf{y}_i\}\big|}{\big|\{v_i \in \mathcal{M}_k : \mathbf{y}_i \neq \emptyset\}\big|}.
\label{eq:leaf-purity}
\end{equation}
LP averages this across leaves:
\begin{equation}
\mathrm{LP} = \frac{1}{K} \sum_{k=1}^{K} \mathrm{Purity}(L_k).
\label{eq:lp}
\end{equation}
LP measures topic homogeneity within each leaf cluster.

\paragraph{PP --- Leaf Path Purity.}
PP applies the same definition but operates over section-path labels $\mathbf{h}_i$ instead of label sets $\mathbf{y}_i$:
\begin{equation}
\mathrm{PP} = \frac{1}{K}\sum_{k=1}^{K} \frac{\displaystyle\max_{h \in \mathcal{H}} \big|\{v_i \in \mathcal{M}_k : h \in \mathbf{h}_i\}\big|}{\big|\{v_i \in \mathcal{M}_k : \mathbf{h}_i \neq \emptyset\}\big|}.
\label{eq:pp}
\end{equation}
PP is stricter than LP: papers grouped in the same leaf must agree not only on labels but on the full hierarchical path leading to them in the survey taxonomy.

\paragraph{NMI --- Normalized Mutual Information.}
To compare predicted trees against flat-clustering baselines under equal granularity, we truncate $\mathcal{T}$ at depth $2$, obtaining a per-paper cluster assignment $\hat{c}_i \in \{1, \ldots, K'\}$. The reference class for paper $v_i$ is its lexicographically smallest label $y_i = \min(\mathbf{y}_i)$. Restricting to papers with $\mathbf{y}_i \neq \emptyset$ and $\hat{c}_i$ defined,
\begin{equation}
\mathrm{NMI} = \frac{2\, I(\hat{C}; Y)}{H(\hat{C}) + H(Y)},
\label{eq:nmi}
\end{equation}
where $I(\cdot; \cdot)$ is mutual information and $H(\cdot)$ is entropy, following the arithmetic-mean normalization in scikit-learn.

\paragraph{CDA --- Citation Direction Accuracy.}
For a citation edge $(v_s, v_d) \in \mathcal{E}_{\mathcal{G}}$ in which $v_s$ cites $v_d$, $v_s$ is typically the more recent work building on $v_d$. We expect $v_s$ to attach at a depth no shallower than $v_d$. To suppress citations within the same year (which carry weak directional information), we restrict to edges with year gap $\tau(v_s) - \tau(v_d) \geq 1$:
\begin{equation}
\small
\begin{aligned}
    \mathrm{CDA} = \\
    \frac{\big|\{(v_s, v_d)\in \mathcal{E}_{\mathcal{G}} : \delta(v_s) \geq \delta(v_d),\; \tau(v_s) - \tau(v_d) \geq 1\}\big|}{\big|\{(v_s, v_d) \in \mathcal{E}_{\mathcal{G}} : \tau(v_s) - \tau(v_d) \geq 1\}\big|}&.
\end{aligned}
\label{eq:cda}
\end{equation}
CDA assesses whether the predicted depth structure aligns with the directionality of citations.

% \paragraph{PMR --- Path Monotonicity Rate.}
% For each concept node $c$, define its nominal time as the mean publication year over papers in its subtree with valid years:
% \begin{equation}
% \bar{\tau}(c) = \frac{1}{|\mathcal{V}(c)|}\!\!\sum_{v_i \in \mathcal{V}(c)}\!\!\tau(v_i).
% \label{eq:nominal-time}
% \end{equation}
% Let $\Pi$ be the set of root-to-leaf paths in $\mathcal{T}$ containing at least two nodes with valid $\bar{\tau}$. A path $\pi = (c_0, c_1, \ldots, c_L) \in \Pi$ is \emph{temporally monotone} if
% \begin{equation}
% \bar{\tau}(c_0) \leq \bar{\tau}(c_1) \leq \cdots \leq \bar{\tau}(c_L).
% \label{eq:pmr-cond}
% \end{equation}
% PMR is the fraction of monotone paths:
% \begin{equation}
% \mathrm{PMR} = \frac{\big|\{\pi \in \Pi : \pi \text{ is temporally monotone}\}\big|}{|\Pi|}.
% \label{eq:pmr}
% \end{equation}
% PMR is more demanding than checking pairwise temporal admissibility along edges: it requires global monotonicity along the entire path, directly testing the ``deeper $\Rightarrow$ more recent'' hypothesis of evolutionary structure.

\paragraph{PMR --- Path Monotonicity Rate.}
PMR evaluates whether root-to-leaf paths reflect a temporally forward-progressing trajectory at the level of the entire concept body, including descendants.
For each concept node $c$, we use the \emph{subtree-mean time} $\bar{\tau}_{\mathcal{T}}(c) = \tfrac{1}{|\mathcal{V}(c)|} \sum_{v_i \in \mathcal{V}(c)} \tau(v_i)$ as the path statistic (Eq.~\ref{eq:tau-defs}).
Let $\Pi$ be the set of root-to-leaf paths in $\mathcal{T}$ containing at least two nodes with valid $\bar{\tau}$.
A path $\pi = (c_0, c_1, \ldots, c_L) \in \Pi$ is temporally monotone if
\begin{equation}
\bar{\tau}_{\mathcal{T}}(c_0) \leq \bar{\tau}_{\mathcal{T}}(c_1) \leq \cdots \leq \bar{\tau}_{\mathcal{T}}(c_L).
\end{equation}
PMR is the fraction of monotone paths:
\begin{equation}
\text{PMR} = \frac{\bigl|\{\pi \in \Pi : \pi \text{ is temporally monotone}\}\bigr|}{|\Pi|}\label{eq:pmr}.
\end{equation}

\subsection{Few-shot metrics on annotated graphs}

\paragraph{FS-CP --- FS Concept Purity.}
FS-CP replicates LP using single-label concepts $g_i$ in place of multi-label section labels:
\begin{equation}
\mathrm{FS\text{-}CP} = \frac{1}{K}\sum_{k=1}^{K} \frac{\displaystyle\max_{g \in \mathcal{G}^{\mathrm{fs}}} \big|\{v_i \in \mathcal{M}_k : g_i = g\}\big|}{|\mathcal{M}_k|}.
\label{eq:FS-CP}
\end{equation}
FS-CP directly measures how well the predicted leaf clusters align with research concepts and is the primary structural-alignment metric in our evaluation.

\paragraph{FS-EDA --- FS Evolutionary Direction Accuracy.}
FS-EDA mirrors CDA but uses annotated evolution edges $\mathcal{E}^{\mathrm{evo}}$, in which $(v_s, v_d)$ indicates that $v_s$ is an evolutionary descendant (typically more recent) of $v_d$:
\begin{equation}
\mathrm{FS\text{-}EDA} = \frac{\big|\{(v_s, v_d) \in \mathcal{E}^{\mathrm{evo}} : \delta(v_s) \geq \delta(v_d)\}\big|}{|\mathcal{E}^{\mathrm{evo}}|}.
\label{eq:FS-EDA}
\end{equation}
Compared with CDA, FS-EDA replaces the noisy citation graph with a clean signal: an edge in $\mathcal{E}^{\mathrm{evo}}$ reflects ``$v_s$ extends/improves $v_d$'', whereas a citation in $\mathcal{E}_{\mathcal{G}}$ may indicate any of comparison, background, or methodological reuse.

\paragraph{FS-Mar --- FS Marginal AUROC.}
Our model produces a per-paper marginality score $h_i \in [0,1]$ (the marginal-head output from Eq.~\ref{eq:l-mar}). With $y_i^{\mathrm{m,fs}} \in \{0,1\}$ as ground truth,
\begin{equation}
\small
\begin{aligned}
    \mathrm{FS\text{-}Mar} = \frac{\big|\{(v_i, v_j) : y_i^{\mathrm{m,fs}}=1,\; y_j^{\mathrm{m,fs}}=0,\; h_i > h_j\}\big|}{\big|\{(v_i, v_j) : y_i^{\mathrm{m,fs}}=1,\; y_j^{\mathrm{m,fs}}=0\}\big|},
\end{aligned}
\label{eq:FS-Mar}
\end{equation}
which equals the standard AUROC of $h$ against $y^{\mathrm{m,fs}}$. Marginal papers are those occupying domain boundaries: relevant by topic, but whose research focus deviates from the core methodological lineage. FS-Mar evaluates a capability unique to our framework---neither standard clustering baselines nor LLM baselines produce a marginality score (reported as ``--'' in Table~\ref{tab:exp2_gt}; a constant score would give the trivial AUROC of $0.500$).

% \newpage
\section{Notation Summary}\label{app:notation}

Table~\ref{tab:notation} summarizes the notation used throughout the paper. Symbols are grouped by category for ease of reference.

\begin{table*}[h]
\centering
\small
\caption{Notation summary. Symbols introduced in the problem formulation (\S\ref{sec:problem}) are used consistently across the Methodology and Appendix.}
\label{tab:notation}
\begin{tabular}{l l l}
\toprule
\textbf{Symbol} & \textbf{Meaning} & \textbf{Introduced} \\
\midrule
\multicolumn{3}{l}{\emph{Input citation graph}} \\
$\mathcal{G} = (\mathcal{V}, \mathcal{E}_{\mathcal{G}})$ & Citation graph from survey reference set & \S\ref{sec:problem} \\
$v_i \in \mathcal{V}$ & A paper node & \S\ref{sec:problem}  \\
$s_i$ & Title and abstract text of paper $v_i$ & \S\ref{sec:problem}  \\
$\tau(v_i)$ & Publication year of paper $v_i$ & \S\ref{sec:problem}  \\
$\mathcal{N}_{\mathcal{G}}(v_i)$ & Citation neighborhood of $v_i$ & App.~\ref{app:encoder} \\
\midrule
\multicolumn{3}{l}{\emph{Trees and concepts}} \\
$\mathcal{T}^{\mathrm{tax}}$ & Time-agnostic taxonomy backbone & \S\ref{sec:problem}  \\
$\mathcal{T}$ & Target evolutionary tree & \S\ref{sec:problem}  \\
$\mathcal{T}^{(t)}, \mathcal{C}^{(t)}$ & Tree and concept set at EM iteration $t$ & \S\ref{sec:3.4.1} \\
$c \in \mathcal{C}$ & Concept node (also indexes a cluster) & \S\ref{sec:problem}  \\
$r$ & Root concept node & \S\ref{sec:3.4.2} \\
$\mathcal{E}_C, \mathcal{E}_P$ & Concept edges, paper-attachment edges & \S\ref{sec:problem}  \\
$\mathcal{V}(c)$ & Papers attached to $c$ or its descendants & \S\ref{sec:problem}  \\
$\mathrm{children}(c)$ & Direct children of $c$ in $\mathcal{T}$ & \S\ref{sec:3.6} \\
$\mathrm{Leaves}(\mathcal{T})$ & Leaf concept nodes of $\mathcal{T}$ & \S\ref{sec:3.4.3} \\
$\phi(c)$ & Concept label of $c$ & \S\ref{sec:problem}  \\
$\tau^{\text{att}}(c), \bar{\tau}(c)$ & Attached-mean / subtree-mean publication year of $c$ & \S\ref{sec:problem} \\
$\epsilon$ & Temporal tolerance margin & \S\ref{sec:problem} \\
\midrule
\multicolumn{3}{l}{\emph{Representations}} \\
$\mathbf{x}_i$ & Pretrained LM embedding of $s_i$ & \S\ref{sec:3.2.1} \\
$\mathbf{z}_i^{\mathrm{sem}}, \mathbf{z}_i^{\mathrm{graph}}\,(=\mathbf{h}_i^{(L)})$ & Semantic / graph stream output & \S\ref{sec:3.2.1} \\
$\mathbf{e}_i$ & Fused clustering embedding & \S\ref{sec:3.2.1} \\
$\mathbf{p}_i$ & Projection representation (contrastive only) & \S\ref{sec:3.2.1} \\
$\boldsymbol{\mu}_c, \boldsymbol{\sigma}_c^2$ & Diagonal Gaussian parameters of concept $c$ & \S\ref{sec:3.2.2} \\
$w_{ic}$ & Soft membership $p(v_i \in C_c)$ & App.~\ref{app:tree-builder} \\
\midrule
\multicolumn{3}{l}{\emph{Marginal mechanism}} \\
$\mathcal{M}$ & Marginal paper set & \S\ref{sec:problem}  \\
$m(v_i, c)$ & Paper-to-concept compatibility score & \S\ref{sec:problem}  \\
$A_i^{\mathrm{leaf}}$ & Maximum leaf compatibility for $v_i$ & \S\ref{sec:3.4.3} \\
$\eta$ & Marginal threshold & \S\ref{sec:problem}  \\
$s_{\mathrm{sem}}, s_{\mathrm{graph}}, s_{\mathrm{time}}$ & Compatibility-score components & App.~\ref{app:marginal} \\
$\gamma_1, \gamma_2, \gamma_3$ & Compatibility-score weights & App.~\ref{app:marginal} \\
\midrule
\multicolumn{3}{l}{\emph{Constraints and constants}} \\

$\rho$ & Major-branch minimum size ratio & \S\ref{sec:3.4.2} \\
$\mathrm{MBC}(\mathcal{T})$ & Major Branch Count of $\mathcal{T}$ & \S\ref{sec:3.4.2} \\
$[d_{\min}, d_{\max}]$ & Drift-window bounds & App.~\ref{app:posttrain_loss} \\
\bottomrule
\end{tabular}
\end{table*}

\newpage
\section{Graph-Aware Paper Encoder}\label{app:encoder}

This appendix expands the abstract encoder $f_{\theta}$ introduced in \S\ref{sec:3.2.1} into concrete forward computations and justifies the three design choices discussed in the main text.

\subsection{Forward computation}

A pretrained scientific encoder (SPECTER2~\cite{Singh2022SciRepEvalAM}) produces a fixed semantic representation
\begin{equation}
\mathbf{x}_i = f_{\mathrm{LM}}(s_i) \in \mathbb{R}^{d_{\mathrm{LM}}},
\label{eq:lm-embed}
\end{equation}
which is then passed through two parallel branches.

\paragraph{Semantic branch.}
A lightweight two-layer MLP produces the semantic stream
\begin{equation}
\mathbf{z}_i^{\mathrm{sem}} = \mathbf{W}_2\,\mathrm{GELU}(\mathbf{W}_1 \mathbf{x}_i),
\label{eq:sem-branch}
\end{equation}
where $\mathbf{W}_1 \in \mathbb{R}^{d \times d_{\mathrm{LM}}}$ and $\mathbf{W}_2 \in \mathbb{R}^{d \times d}$.

\paragraph{Graph branch.}
A stack of $L$ controlled propagation blocks computes the graph stream. With $\mathbf{h}_i^{(0)} = \mathrm{MLP}_{\mathrm{graph}}(\mathbf{x}_i)$, each layer $l = 0, \ldots, L-1$ updates
\begin{align}
\mathbf{m}_i^{(l)} &= \frac{1}{|\mathcal{N}_{\mathcal{G}}(v_i)|}\!\!\sum_{j \in \mathcal{N}_{\mathcal{G}}(v_i)}\!\! \mathbf{W}_{\mathrm{msg}}\, \mathbf{h}_j^{(l)}, \label{eq:msg}\\
\mathbf{u}_i^{(l)} &= \beta\, \mathbf{h}_i^{(l)} + (1-\beta)\, \mathbf{m}_i^{(l)}, \label{eq:mix}\\
\mathbf{h}_i^{(l+1)} &= \mathrm{PairNorm}\!\left(\alpha\, \mathbf{x}_i + (1-\alpha)\, \mathbf{u}_i^{(l)}\right). \label{eq:propagation}
\end{align}
For isolated nodes $v_i \in \mathcal{V}_o$ where $|\mathcal{N}_{\mathcal{G}}(v_i)| = 0$, neighborhood aggregation is undefined; we set $\mathbf{m}_i^{(l)} = \mathbf{0}$ so that the propagation reduces to $\mathbf{h}_i^{(l+1)} = \mathrm{PairNorm}(\alpha \mathbf{x}_i + (1-\alpha)\beta \mathbf{h}_i^{(l)})$, i.e., a pure semantic residual. The final graph-stream output is $\mathbf{z}_i^{\mathrm{graph}} = \mathbf{h}_i^{(L)}$.

\paragraph{Adaptive fusion gate.}
The two streams are combined through a learnable connectivity-biased gate:
\begin{align}
\lambda_i &= \sigma\!\left(\mathbf{W}_g [\mathbf{z}_i^{\mathrm{graph}} \,\|\, \mathbf{z}_i^{\mathrm{sem}}] + \delta\, \mathbb{1}[\mathrm{iso}(v_i)]\right), \label{eq:gate}\\
\mathbf{z}_i^{\mathrm{fused}} &= \lambda_i\, \mathbf{z}_i^{\mathrm{graph}} + (1-\lambda_i)\, \mathbf{z}_i^{\mathrm{sem}}. \label{eq:fused}
\end{align}
The clustering embedding is $\mathbf{e}_i = \mathrm{L_2Norm}(\mathbf{z}_i^{\mathrm{fused}})$. An independent two-layer MLP produces the projection representation $\mathbf{p}_i = \mathrm{MLP}_{\mathrm{proj}}(\mathbf{z}_i^{\mathrm{fused}})$, used exclusively by the contrastive objective in \S\ref{sec:3.3} and never as input to clustering or downstream attachment.

The three design choices flagged in the main text correspond to specific equations above:
\begin{itemize}[leftmargin=*,topsep=2pt,itemsep=1pt]
\item \textbf{Anchored residual:} the $\alpha \mathbf{x}_i$ term in Eq.~\ref{eq:propagation} preserves the original semantic vector at every layer, preventing the propagation from collapsing distinct papers to similar representations in shallow citation graphs.
\item \textbf{Density-aware normalization:} $\mathrm{PairNorm}$ in Eq.~\ref{eq:propagation} centers features and rescales pairwise distances, accommodating the highly uneven cluster densities induced by hub papers.
\item \textbf{Connectivity-biased fusion gate:} the term $\delta \mathbb{1}[\mathrm{iso}(v_i)]$ in Eq.~\ref{eq:gate}, with $\delta$ initialized to $-2.0$, drives $\lambda_i$ toward $0$ for isolated nodes so that $\mathbf{z}_i^{\mathrm{fused}} \approx \mathbf{z}_i^{\mathrm{sem}}$ at initialization, while remaining learnable.
\end{itemize}

% \subsection{Hyperparameters}

% \begin{itemize}[leftmargin=*,topsep=2pt,itemsep=1pt]
% \item Pretrained encoder: SPECTER2, $d_{\mathrm{LM}} = 768$.
% \item Hidden dimension $d = 256$, propagation layers $L = 2$.
% \item Anchor coefficient $\alpha = 0.3$, mixing coefficient $\beta = 0.5$, both learnable.
% \item Isolated-node bias $\delta$: learnable scalar, initialized to $-2.0$.
% \item PairNorm: scale $1.0$, applied after every propagation step.
% \item Activation: GELU; dropout $0.1$ between MLP layers.
% \item Projection head: two-layer MLP $256 \to 256 \to 128$.
% \end{itemize}

\newpage
\section{Distributional Tree Builder}\label{app:tree-builder}

\subsection{Base clustering with soft membership}

HDBSCAN is applied to the L2-normalized embeddings $\{\mathbf{e}_i\}$ to identify dense regions. To obtain soft membership, we use the algorithm's native extension based on the mutual reachability tree~\citep{campello2013density}, which yields
\begin{equation}
w_{ic} = p(v_i \in C_c) \in [0, 1], \quad c = 1, \ldots, K,
\label{eq:soft-membership}
\end{equation}
where $K$ is the number of discovered base clusters. Each cluster's diagonal Gaussian is fitted by weighted maximum likelihood:
\begin{align}
\boldsymbol{\mu}_c &= \frac{\sum_i w_{ic}\, \mathbf{e}_i}{\sum_i w_{ic}}, \label{eq:weighted-mean}\\
\sigma_{c,d}^2 &= \max\!\left(\frac{\sum_i w_{ic}\,(e_{i,d} - \mu_{c,d})^2}{\sum_i w_{ic}},\; \sigma_{\min}^2\right), \label{eq:weighted-var}
\end{align}
where $d$ indexes embedding dimensions and $\sigma_{\min}^2$ is a numerical floor preventing degenerate Gaussians.

\subsection{2-Wasserstein distance between diagonal Gaussians}

For two diagonal Gaussians $C_p \sim \mathcal{N}(\boldsymbol{\mu}_p, \mathrm{diag}(\boldsymbol{\sigma}_p^2))$ and $C_q \sim \mathcal{N}(\boldsymbol{\mu}_q, \mathrm{diag}(\boldsymbol{\sigma}_q^2))$, the squared 2-Wasserstein distance has a closed form:
\begin{equation}
\mathcal{W}_2^2(C_p, C_q) = \|\boldsymbol{\mu}_p - \boldsymbol{\mu}_q\|_2^2 + \sum_d (\sigma_{p,d} - \sigma_{q,d})^2.
\label{eq:w2-closed}
\end{equation}

\subsection{Bottom-up agglomeration}

Starting from the $K$ base clusters, we greedily merge the pair $(C_p^\star, C_q^\star) = \arg\min_{p \neq q} \mathcal{W}_2(C_p, C_q)$ at each step. The merged parent's Gaussian is re-estimated from the union of member papers $M = \mathrm{members}(C_p^\star) \cup \mathrm{members}(C_q^\star)$ via unweighted maximum likelihood:
\begin{equation}
\begin{aligned}
\boldsymbol{\mu}_{\mathrm{parent}} &= \frac{1}{|M|}\sum_{i \in M} \mathbf{e}_i,\\
\sigma_{\mathrm{parent},d}^2 &= \\\max&\!\left(\frac{1}{|M|}\sum_{i \in M}(e_{i,d} - \mu_{\mathrm{parent},d})^2,\; \sigma_{\min}^2\right). 
\end{aligned}
\label{eq:parent-var}
\end{equation}
Re-estimation relies only on running sums $\sum_{i \in M} \mathbf{e}_i$ and sum of squares $\sum_{i \in M} \mathbf{e}_i^{\odot 2}$, which can be incrementally maintained as sufficient statistics during agglomeration. The procedure terminates when a single root cluster remains. Algorithm~\ref{alg:tree-builder} summarizes the full builder.

\begin{algorithm}[h]
\caption{Distributional tree builder.}
\label{alg:tree-builder}
\small
\begin{algorithmic}[1]
\REQUIRE Embeddings $\{\mathbf{e}_i\}$; HDBSCAN parameters; (optional) temporal admissibility filter $\mathcal{F}_\epsilon$.
\STATE $\{w_{ic}\}, K \gets \mathrm{HDBSCAN\_SoftMembership}(\{\mathbf{e}_i\})$
\FOR{$c = 1, \ldots, K$}
    \STATE Compute $\boldsymbol{\mu}_c, \boldsymbol{\sigma}_c^2$ via Eq.~\ref{eq:weighted-mean}--\ref{eq:weighted-var}.
\ENDFOR
\STATE $\mathcal{C} \gets \{C_1, \ldots, C_K\}$; record member papers per cluster.
\WHILE{$|\mathcal{C}| > 1$}
    \STATE Compute $\mathcal{W}_2(C_p, C_q)$ for all pairs (Eq.~\ref{eq:w2-closed}).
    \STATE Select $(C_p^\star, C_q^\star) \gets \arg\min \mathcal{W}_2$ \textbf{satisfying $\mathcal{F}_\epsilon$ if provided}.
    \STATE Re-estimate parent via Eq.~\ref{eq:parent-var}; remove children, add parent.
\ENDWHILE
\RETURN Hierarchical tree with $(\boldsymbol{\mu}_c, \boldsymbol{\sigma}_c^2)$ at every node.
\end{algorithmic}
\end{algorithm}

% \subsection{Hyperparameters}

% \begin{itemize}[leftmargin=*,topsep=2pt,itemsep=1pt]
% \item HDBSCAN: \texttt{min\_cluster\_size}~=~5, \texttt{min\_samples}~=~3, metric~=~cosine, \texttt{cluster\_selection\_epsilon}~=~0.0.
% \item Variance floor $\sigma_{\min}^2 = 10^{-4}$.
% \item Soft-membership extension via \texttt{all\_points\_membership\_vectors}.
% \end{itemize}

\section{Stage I Loss Details}\label{app:pretrain_loss}

This appendix expands the three pre-training loss groups summarized in \S\ref{sec:3.3}:
\begin{equation*}
\mathcal{L}^{\mathrm{tax}} = \mathcal{L}_{\mathrm{sec}} + \mathcal{L}_{\mathrm{cite}} + \mathcal{L}_{\mathrm{geo}}.
\end{equation*}

\subsection{Section supervision \texorpdfstring{$\mathcal{L}_{\mathrm{sec}}$}{L\_sec}}

Survey section structure supplies weak labels: each paper $v_i$ inherits a multi-hot section vector $\mathbf{y}_i^{\mathrm{lbl}} \in \{0,1\}^{|\mathcal{S}|}$ over the union of section labels $\mathcal{S}$, plus a hierarchical level $y_i^{\mathrm{lvl}} \in \{1, \ldots, L_{\max}\}$. Let $\hat{\mathbf{y}}_i^{\mathrm{lbl}} = \mathrm{sigmoid}(\mathbf{W}_{\mathrm{lbl}} \mathbf{e}_i)$ and $\hat{\mathbf{y}}_i^{\mathrm{lvl}} = \mathrm{softmax}(\mathbf{W}_{\mathrm{lvl}} \mathbf{e}_i)$. We use masked binary cross-entropy and categorical cross-entropy over labeled nodes $\mathcal{V}_{\mathrm{lbl}} \subseteq \mathcal{V}$:
\begin{align}
\mathcal{L}_{\mathrm{label}} &= -\frac{1}{|\mathcal{V}_{\mathrm{lbl}}|}\!\!\sum_{i \in \mathcal{V}_{\mathrm{lbl}}}\!\! \mathrm{BCE}\!\left(\hat{\mathbf{y}}_i^{\mathrm{lbl}}, \mathbf{y}_i^{\mathrm{lbl}}\right), \label{eq:l-label}\\
\mathcal{L}_{\mathrm{level}} &= -\frac{1}{|\mathcal{V}_{\mathrm{lbl}}|}\!\!\sum_{i \in \mathcal{V}_{\mathrm{lbl}}}\!\! \log \hat{y}_{i, y_i^{\mathrm{lvl}}}^{\mathrm{lvl}}, \label{eq:l-level}\\
\mathcal{L}_{\mathrm{sec}} &= w_{\mathrm{lbl}}\mathcal{L}_{\mathrm{label}} + w_{\mathrm{lvl}}\mathcal{L}_{\mathrm{level}}.
\end{align}

\subsection{Citation consistency \texorpdfstring{$\mathcal{L}_{\mathrm{cite}}$}{L\_cite}}

Two terms preserve citation structure: link prediction on $\mathbf{e}_i$ and contrastive ranking on $\mathbf{p}_i$.

\paragraph{Link prediction.}
For each observed edge $(v_i, v_j) \in \mathcal{E}_{\mathcal{G}}$ (positive) and randomly sampled non-edge $(v_i, v_{j^-})$ (negative), define the logit $\hat{y}_{ij} = \tau_{\mathrm{lnk}}\, \mathbf{e}_i^\top \mathbf{e}_j$ with temperature $\tau_{\mathrm{lnk}}$ (we use $10$ to scale cosine similarities into a useful logit range):
\begin{equation}
\mathcal{L}_{\mathrm{link}} = \mathrm{BCE}\!\left(\mathrm{sigmoid}(\hat{y}_{ij}),\, y_{ij}\right),
\label{eq:l-link}
\end{equation}
where $y_{ij} \in \{0,1\}$ indicates edge existence. We use one negative per positive.

\paragraph{Contrastive ranking on $\mathbf{p}_i$.}
For two papers $v_i, v_j$ sharing a section label, they form a positive pair. Hard negatives $v_{k^-}$ are mined as the most similar paper to $v_i$ not sharing any label of $v_i$. With cosine similarity $\mathrm{sim}(a, b) = \mathbf{p}_a^\top \mathbf{p}_b / (\|\mathbf{p}_a\|\|\mathbf{p}_b\|)$ and temperature $\tau_{\mathrm{ctr}}$:
\begin{equation}
\begin{aligned}
    \mathcal{L}_{\mathrm{contrast}} =\\
    -\log &\frac{\exp(\mathrm{sim}(v_i, v_j)/\tau_{\mathrm{ctr}})}{\sum_{k \in \{j\} \cup \mathcal{K}^-_i} \exp(\mathrm{sim}(v_i, v_k)/\tau_{\mathrm{ctr}})},
\end{aligned}
\label{eq:l-contrast}
\end{equation}
with $\mathcal{K}^-_i$ the hard-negative set ($|\mathcal{K}^-_i| = 8$). Note that $\mathcal{L}_{\mathrm{contrast}}$ acts on $\mathbf{p}_i$ and never on $\mathbf{e}_i$, keeping the clustering geometry untouched.

\paragraph{Combination.}
\begin{equation}
\mathcal{L}_{\mathrm{cite}} = w_{\mathrm{lnk}}\mathcal{L}_{\mathrm{link}} + w_{\mathrm{ctr}}\mathcal{L}_{\mathrm{contrast}}.
\end{equation}

\subsection{Cluster geometry \texorpdfstring{$\mathcal{L}_{\mathrm{geo}}$}{L\_geo}}

A prototype-anchoring term pulls each labeled node toward its class centroid. For each label $s \in \mathcal{S}$, let $\mathcal{V}_s = \{v_i : s \in \mathbf{y}_i^{\mathrm{lbl}}\}$ and define the stop-gradient centroid
\begin{equation}
\bar{\mathbf{e}}_s = \mathrm{sg}\!\left(\frac{1}{|\mathcal{V}_s|}\!\sum_{i \in \mathcal{V}_s}\!\! \mathbf{e}_i\right).
\label{eq:proto-centroid}
\end{equation}
The anchor loss is:
\begin{equation}
\mathcal{L}_{\mathrm{anchor}} = \frac{1}{|\mathcal{S}|}\!\sum_{s \in \mathcal{S}}\!\frac{1}{|\mathcal{V}_s|}\!\sum_{i \in \mathcal{V}_s}\!\!\left(1 - \mathbf{e}_i^\top \bar{\mathbf{e}}_s\right).
\label{eq:l-anchor}
\end{equation}
$\mathcal{L}_{\mathrm{geo}} = w_{\mathrm{anc}}\mathcal{L}_{\mathrm{anchor}}$.

% \subsection{Weights}

% \begin{table}[h]
% \centering
% \small
% \begin{tabular}{l c l}
% \toprule
% Component & Weight & Eq. \\
% \midrule
% $\mathcal{L}_{\mathrm{label}}$ & $w_{\mathrm{lbl}} = 1.5$ & Eq.~\ref{eq:l-label} \\
% $\mathcal{L}_{\mathrm{level}}$ & $w_{\mathrm{lvl}} = 0.9$ & Eq.~\ref{eq:l-level} \\
% $\mathcal{L}_{\mathrm{link}}$ & $w_{\mathrm{lnk}} = 0.1$ & Eq.~\ref{eq:l-link} \\
% $\mathcal{L}_{\mathrm{contrast}}$ & $w_{\mathrm{ctr}} = 0.3$ & Eq.~\ref{eq:l-contrast} \\
% $\mathcal{L}_{\mathrm{anchor}}$ & $w_{\mathrm{anc}} = 0.5$ & Eq.~\ref{eq:l-anchor} \\
% \bottomrule
% \end{tabular}
% \end{table}

% Temperatures: $\tau_{\mathrm{lnk}} = 10$, $\tau_{\mathrm{ctr}} = 0.07$.

\section{Stage II Loss Details}\label{app:posttrain_loss}

This appendix expands the three evolution loss groups summarized in \S\ref{sec:3.4.1}:
\begin{equation*}
\mathcal{L}^{\mathrm{evo}} = \mathcal{L}_{\mathrm{sem}} + \mathcal{L}_{\mathrm{temp}} + \mathcal{L}_{\mathrm{marg}}.
\end{equation*}
All terms operate on the current tree $\mathcal{T}^{(t)}$ with concept Gaussians $\{(\boldsymbol{\mu}_c, \boldsymbol{\sigma}_c^2)\}_{c \in \mathcal{C}^{(t)}}$. Let $p(c)$ denote the parent of concept $c$ and $p(i)$ the parent concept of paper $v_i$ under attachment $a$.

\subsection{Semantic evolution consistency \texorpdfstring{$\mathcal{L}_{\mathrm{sem}}$}{L\_sem}}

Three sub-terms enforce parent--child semantic coherence.

\paragraph{Node-to-parent pull.}
Each paper is pulled toward the centroid of its parent concept:
\begin{equation}
\mathcal{L}_{\mathrm{sem\text{-}n}} = \frac{1}{|\mathcal{V}|}\sum_{i \in \mathcal{V}}\!\left(1 - \cos(\mathbf{e}_i, \boldsymbol{\mu}_{p(i)})\right).
\label{eq:l-semn}
\end{equation}

\paragraph{Cluster-to-parent pull.}
Each concept is pulled toward the centroid of its parent concept:
\begin{equation}
\begin{aligned}
    \mathcal{L}_{\mathrm{sem\text{-}c}} = \\\frac{1}{|\mathcal{C}^{(t)} \setminus \{r\}|}&\!\!\sum_{c \in \mathcal{C}^{(t)} \setminus \{r\}}\!\!\!\left(1 - \cos(\boldsymbol{\mu}_c, \boldsymbol{\mu}_{p(c)})\right).
\end{aligned}
\label{eq:l-semc}
\end{equation}

\paragraph{Drift window.}
Let $\delta_c = \|\boldsymbol{\mu}_c - \boldsymbol{\mu}_{p(c)}\|_2$. A two-sided hinge confines $\delta_c$ to $[d_{\min}, d_{\max}]$:
\begin{equation}
\small
\begin{aligned}
    \mathcal{L}_{\mathrm{drift}} = \\\!\!\sum_{c \in \mathcal{C}^{(t)} \setminus \{r\}}&\!\!\!\big[\max(0, d_{\min} - \delta_c) + \max(0, \delta_c - d_{\max})\big].
\end{aligned}
\label{eq:l-drift}
\end{equation}
% We use $[d_{\min}, d_{\max}] = [0.05, 0.50]$.

\paragraph{Combination.}
\begin{equation}
\mathcal{L}_{\mathrm{sem}} = w_{\mathrm{semn}}\mathcal{L}_{\mathrm{sem\text{-}n}} + w_{\mathrm{semc}}\mathcal{L}_{\mathrm{sem\text{-}c}} + w_{\mathrm{drift}}\mathcal{L}_{\mathrm{drift}}.
\end{equation}

\subsection{Temporal coherence \texorpdfstring{$\mathcal{L}_{\mathrm{temp}}$}{L\_temp}}

Three sub-terms enforce forward-progressing evolution.

\paragraph{Path smoothness.}
For a root-to-leaf concept path $P = (c_0, c_1, \ldots, c_L)$, define velocity $\Delta_l = \boldsymbol{\mu}_{c_l} - \boldsymbol{\mu}_{c_{l-1}}$. We penalize second-order changes:
\begin{equation}
\mathcal{L}_{\mathrm{smooth}} = \frac{1}{|\mathcal{P}|}\!\sum_{P \in \mathcal{P}}\!\frac{1}{L-1}\!\sum_{l=2}^{L}\!\|\Delta_l - \Delta_{l-1}\|_2^2,
\label{eq:l-smooth}
\end{equation}
where $\mathcal{P}$ is the set of root-to-leaf paths in $\mathcal{T}^{(t)}$.

\paragraph{Path monotonicity.}
Let $\tau_T(c) = \min_{v \in \mathcal{V}(c)} \tau(v)$ denote the earliest publication year in the subtree of $c$ (also used in Appendix~\ref{app:constraints}). A hinge encourages child concepts to be no earlier than parents:
\begin{equation}
\mathcal{L}_{\mathrm{path\text{-}mono}} = \!\!\!\!\!\sum_{(c_p, c_c) \in \mathcal{E}_C}\!\!\!\!\!\max(0,\, \tau_T(c_p) - \tau_T(c_c) - \epsilon).
\label{eq:l-pathmono}
\end{equation}

\paragraph{Citation directionality.}
For each citation $(v_i, v_j) \in \mathcal{E}_{\mathcal{G}}$ with $v_i$ citing $v_j$, the citing paper should attach to a deeper node. Let $q_{i,c}$ be the soft attachment probability of $v_i$ to concept $c$:
\begin{equation}
q_{i,c} = \frac{\exp\!\left(-\|\mathbf{e}_i - \boldsymbol{\mu}_c\|_2^2 / \tau_{\mathrm{att}}\right)}{\sum_{c' \in \mathcal{C}^{(t)}}\!\exp\!\left(-\|\mathbf{e}_i - \boldsymbol{\mu}_{c'}\|_2^2 / \tau_{\mathrm{att}}\right)}.
\label{eq:soft-att}
\end{equation}
Define the expected depth $\bar{d}(v_i) = \sum_{c} q_{i,c}\, \mathrm{depth}(c)$. The citation directionality loss is:
\begin{equation}
\mathcal{L}_{\mathrm{cite\text{-}dir}} = \!\!\!\!\!\sum_{(v_i, v_j) \in \mathcal{E}_{\mathcal{G}}}\!\!\!\!\!\max(0,\, \bar{d}(v_j) - \bar{d}(v_i) + \xi),
\label{eq:l-citedir}
\end{equation}
penalizing cases where the cited paper $v_j$ ends up at least $\xi$ levels deeper than the citing $v_i$.

\paragraph{Combination.}
\begin{equation}
\mathcal{L}_{\mathrm{temp}} = \mathcal{L}_{\mathrm{smooth}} + \mathcal{L}_{\mathrm{path\text{-}mono}} + \mathcal{L}_{\mathrm{cite\text{-}dir}}.
\end{equation}

\subsection{Marginal identification and re-attachment \texorpdfstring{$\mathcal{L}_{\mathrm{marg}}$}{L\_marg}}

Let $h_i = \mathrm{MLP}_{\mathrm{mar}}(\mathbf{e}_i) \in [0,1]$ be the marginal-head output. Pseudo-labels are $\tilde{y}_i^{\mathrm{m}} = \mathbb{1}[A_i^{\mathrm{leaf}} < \eta]$.

\paragraph{Marginal classification.}
\begin{equation}
\small
\mathcal{L}_{\mathrm{mar}} = -\frac{1}{|\mathcal{V}|}\!\sum_{i \in \mathcal{V}}\!\big[\tilde{y}_i^{\mathrm{m}} \log h_i + (1-\tilde{y}_i^{\mathrm{m}}) \log(1-h_i)\big].
\label{eq:l-mar}
\end{equation}

\paragraph{Attachment ranking.}
For each flagged marginal paper $v_m \in \mathcal{M}^{(t)}$, let $c^+_m = a(v_m)$ (the optimal attachment from Eq.~\ref{eq:marginal-attach}) and $c^-$ a candidate concept that fails the temporal admissibility check. With score
\begin{equation*}
s_{\mathrm{att}}(c, m) = \log \mathcal{N}\!\left(\mathbf{e}_m;\, \boldsymbol{\mu}_c, \mathrm{diag}(\boldsymbol{\sigma}_c^2)\right),
\end{equation*}
\begin{equation}
\small
\begin{aligned}
\mathcal{L}_{\mathrm{att}}^{\mathrm{rank}} = \\
\!\!\sum_{v_m \in \mathcal{M}^{(t)}}\!\!\!\sum_{c^-}\! &\max\!\big(0,\, s_{\mathrm{att}}(c^-, m) - s_{\mathrm{att}}(c^+_m, m) + \zeta\big),
\end{aligned}
\label{eq:l-attrank}
\end{equation}
with margin $\zeta$.

\paragraph{Centroid repulsion.}
To prevent internal concept centroids from collapsing, sibling concepts are encouraged to maintain pairwise separation. For each internal concept $c$ with children $\mathrm{children}(c)$:
\begin{equation}
\begin{aligned}
    \mathcal{L}_{\mathrm{repulsion}} = \\\!\!\sum_{c}\!\!\sum_{\substack{c_a, c_b \in \mathrm{children}(c) \\ a \neq b}}\!\!\!\!&\max\!\big(0,\, \zeta_r - \|\boldsymbol{\mu}_{c_a} - \boldsymbol{\mu}_{c_b}\|_2\big),
\end{aligned}
\label{eq:l-repulsion}
\end{equation}
with separation margin $\zeta_r$.

\paragraph{Combination.}
\begin{equation}
\mathcal{L}_{\mathrm{marg}} = \mathcal{L}_{\mathrm{mar}} + \mathcal{L}_{\mathrm{att}}^{\mathrm{rank}} + \mathcal{L}_{\mathrm{repulsion}}.
\end{equation}

% \subsection{Auxiliary link signal}

% The pre-training $\mathcal{L}_{\mathrm{link}}$ term (Eq.~\ref{eq:l-link}) is retained at a small weight throughout Stage II to prevent citation structure from drifting under the new objectives.

% \subsection{Weights}

% \begin{table}[h]
% \centering
% \small
% \begin{tabular}{l l c}
% \toprule
% Group & Term & Weight \\
% \midrule
% \multirow{3}{*}{Semantic} & $\mathcal{L}_{\mathrm{sem\text{-}n}}$ & $w_{\mathrm{semn}} = 0.3$ \\
%  & $\mathcal{L}_{\mathrm{sem\text{-}c}}$ & $w_{\mathrm{semc}} = 0.3$ \\
%  & $\mathcal{L}_{\mathrm{drift}}$ & $w_{\mathrm{drift}} = 0.5$ \\
% \midrule
% \multirow{3}{*}{Temporal} & $\mathcal{L}_{\mathrm{smooth}}$ & $w_{\mathrm{sm}} = 0.05$ \\
%  & $\mathcal{L}_{\mathrm{path\text{-}mono}}$ & $w_{\mathrm{pm}} = 0.3$ \\
%  & $\mathcal{L}_{\mathrm{cite\text{-}dir}}$ & $w_{\mathrm{cd}} = 0.3$ \\
% \midrule
% \multirow{3}{*}{Marginal} & $\mathcal{L}_{\mathrm{mar}}$ & $w_{\mathrm{m}} = 0.3$ \\
%  & $\mathcal{L}_{\mathrm{att}}^{\mathrm{rank}}$ & $w_{\mathrm{ar}} = 0.4$ \\
%  & $\mathcal{L}_{\mathrm{repulsion}}$ & $w_{\mathrm{rp}} = 0.5$ \\
% \midrule
% Auxiliary & $\mathcal{L}_{\mathrm{link}}$ & $w_{\mathrm{lnk}} = 0.1$ \\
% \bottomrule
% \end{tabular}
% \end{table}

% % Other hyperparameters: temperature $\tau_{\mathrm{att}} = 0.1$, citation slack $\xi = 0.5$, ranking margin $\zeta = 0.5$, repulsion margin $\zeta_r = 0.2$.

% \newpage
\section{Structural Constraints}\label{app:constraints}

The main text describes the three most consequential constraints (temporal consistency, backbone preservation, tree legality). Here we formalize the two remaining constraints deferred from \S\ref{sec:3.4.2}.

\subsection{Path temporal consistency}

For every root-to-leaf concept path $(c_0, c_1, \ldots, c_L)$ in $\mathcal{T}$,
\begin{equation}
\tau_T(c_0) \leq \tau_T(c_1) + \epsilon \leq \cdots \leq \tau_T(c_L) + L\epsilon,
\label{eq:path-temp}
\end{equation}
where $\tau_T(c) = \min_{v \in \mathcal{V}(c)} \tau(v)$ and $\epsilon$ is the per-edge tolerance margin. This rules out global temporal inversions in which local pairs satisfy the temporal order but the path as a whole accumulates a violation.

\subsection{Path membership coherence}

For every concept $c \in \mathcal{C}$, its evolutionary ancestor chain $\mathrm{anc}^{\mathrm{evo}}(c) = (c, p(c), p(p(c)), \ldots, r)$ must retain non-trivial distributional overlap with its taxonomy-ancestor chain $\mathrm{anc}^{\mathrm{tax}}(c)$. Formally, defining overlap as the maximum normalized intersection between any ancestor in the two chains,
\begin{equation}
\small
\!\!\!\Omega(c) = \max_{c_e \in \mathrm{anc}^{\mathrm{evo}}(c),\, c_t \in \mathrm{anc}^{\mathrm{tax}}(c)}\frac{|\mathcal{V}(c_e) \cap \mathcal{V}(c_t)|}{\min(|\mathcal{V}(c_e)|, |\mathcal{V}(c_t)|)},
\label{eq:path-mem}
\end{equation}
and requiring $\Omega(c) \geq \theta_{\mathrm{ov}}$. We use $\theta_{\mathrm{ov}} = 0.3$. This constraint prevents the temporal reorientation in Stage II from rewiring a concept into a semantically unrelated branch.

% \subsection{Tolerance values}

% $\epsilon = 1$ year (per-edge), $\rho = 0.05$ (major-branch ratio), $\theta_{\mathrm{ov}} = 0.3$ (overlap threshold).

\section{Marginal Compatibility Score}\label{app:marginal}

This appendix specifies the compatibility score $m(v_i, c)$ in \S\ref{sec:3.4.3}.

For paper $v_i$ and concept $c$ with Gaussian $(\boldsymbol{\mu}_c, \boldsymbol{\sigma}_c^2)$ and member set $\mathcal{V}(c)$:

\paragraph{Semantic similarity.}
\begin{equation}
s_{\mathrm{sem}}(i, c) = \frac{1}{2}\!\left(1 + \cos(\mathbf{e}_i, \boldsymbol{\mu}_c)\right) \in [0, 1].
\label{eq:s-sem}
\end{equation}

\paragraph{Citation-neighborhood overlap.}
\begin{equation}
s_{\mathrm{graph}}(i, c) = \frac{|\mathcal{N}_{\mathcal{G}}(v_i) \cap \mathcal{V}(c)|}{\max(|\mathcal{N}_{\mathcal{G}}(v_i)|, 1)} \in [0, 1].
\label{eq:s-graph}
\end{equation}

\paragraph{Temporal compatibility.}
Let $\bar{\tau}_c = |\mathcal{V}(c)|^{-1}\!\sum_{v \in \mathcal{V}(c)} \tau(v)$ be the nominal time of $c$. With time-scale $\sigma_t$:
\begin{equation}
s_{\mathrm{time}}(i, c) = \exp\!\left(-\frac{|\tau(v_i) - \bar{\tau}_c|}{\sigma_t}\right) \in (0, 1].
\label{eq:s-time}
\end{equation}

\paragraph{Combined compatibility.}
\begin{equation}
\begin{aligned}
m(v_i, c) = \\
\gamma_1\, s_{\mathrm{sem}}(i, c) &+ \gamma_2\, s_{\mathrm{graph}}(i, c) + \gamma_3\, s_{\mathrm{time}}(i, c),
\end{aligned}
\label{eq:m-combined}
\end{equation}
with $\gamma_1 + \gamma_2 + \gamma_3 = 1$ and $m(v_i, c) \in [0, 1]$.

% \paragraph{Hyperparameters.}
% $\gamma_1 = 0.5$, $\gamma_2 = 0.3$, $\gamma_3 = 0.2$, $\sigma_t = 3$ years, marginal threshold $\eta = 0.4$.

\section{Stage III Calibration Loss Details}\label{app:gt_loss}

This appendix expands the three calibration loss terms summarized in \S\ref{sec:3.5}:
\begin{equation*}
\mathcal{L}^{\mathrm{cal}} = \mathcal{L}^{\mathrm{cal}}_{\mathrm{marg}} + \mathcal{L}^{\mathrm{cal}}_{\mathrm{concept}} + \mathcal{L}^{\mathrm{cal}}_{\mathrm{edge}}.
\end{equation*}
The FS dataset (\S\ref{sec:datasets}) provides per-paper labels $y_i^{\mathrm{m,fs}} \in \{0,1\}$ (marginal), $g_i \in \mathcal{G}^{\mathrm{fs}}$ (concept), and the evolution edge set $\mathcal{E}^{\mathrm{evo}}$ over papers (Appendix~\ref{app:metrics}).

\subsection{FS marginal classification}

Class-weighted BCE to handle the imbalance between core and marginal papers:
\begin{equation}
\small
\begin{aligned}
\mathcal{L}^{\mathrm{cal}}_{\mathrm{marg}} = -\frac{1}{|\mathcal{V}^{\mathrm{fs}}|}\!\!\sum_{i \in \mathcal{V}^{\mathrm{fs}}}&\big[w^+ y_i^{\mathrm{m,fs}} \log h_i \\
&\quad + w^- (1-y_i^{\mathrm{m,fs}}) \log(1-h_i)\big],
\end{aligned}
\label{eq:l-fsmar}
\end{equation}
with $w^+ = (1-\bar{y}^{\mathrm{m}})/\bar{y}^{\mathrm{m}}$, $w^- = 1$, and $\bar{y}^{\mathrm{m}}$ the marginal fraction in the FS set.

\subsection{FS concept alignment}

For each FS concept $g \in \mathcal{G}^{\mathrm{fs}}$, define its empirical centroid
\begin{equation}
\bar{\mathbf{e}}_g = \mathrm{sg}\!\left(\frac{1}{|\mathcal{V}_g^{\mathrm{fs}}|}\!\!\sum_{i \in \mathcal{V}_g^{\mathrm{fs}}}\!\! \mathbf{e}_i\right),
\label{eq:fs-centroid}
\end{equation}
with $\mathcal{V}_g^{\mathrm{fs}} = \{v_i : g_i = g\}$. An InfoNCE objective aligns each paper with its FS centroid:
\begin{equation}
\small
\mathcal{L}^{\mathrm{cal}}_{\mathrm{concept}} = -\frac{1}{|\mathcal{V}^{\mathrm{fs}}|}\!\!\sum_{i \in \mathcal{V}^{\mathrm{fs}}}\!\!\log\frac{\exp(\mathbf{e}_i^\top \bar{\mathbf{e}}_{g_i} / \tau_{\mathrm{ic}})}{\sum_{g' \in \mathcal{G}^{\mathrm{fs}}}\!\exp(\mathbf{e}_i^\top \bar{\mathbf{e}}_{g'} / \tau_{\mathrm{ic}})}.
\label{eq:l-fsconcept}
\end{equation}

\subsection{FS edge directionality}

For each annotated edge $(v_s, v_d) \in \mathcal{E}^{\mathrm{evo}}$, $v_s$ is the evolutionary descendant of $v_d$. A triplet-margin loss requires the predicted depth of $v_s$ to exceed that of $v_d$, with cross-domain \texttt{adapts} relations receiving a smaller margin:
\begin{equation}
\mathcal{L}^{\mathrm{cal}}_{\mathrm{edge}} = \!\!\!\!\!\!\!\sum_{(v_s, v_d) \in \mathcal{E}^{\mathrm{evo}}}\!\!\!\!\!\!\!\max\!\big(0,\, \bar{d}(v_d) - \bar{d}(v_s) + \mu(v_s, v_d)\big),
\label{eq:l-gtedge}
\end{equation}
where $\bar{d}(\cdot)$ is the expected depth from Eq.~\ref{eq:soft-att}, and
\begin{equation*}
\small
\mu(v_s, v_d) = \begin{cases} \mu_{\mathrm{strict}} & \text{\texttt{extends} / \texttt{improves}}, \\ \mu_{\mathrm{relax}} & \text{\texttt{adapts}}. \end{cases}
\end{equation*}

% \subsection{Weights and learning rate}

% \begin{itemize}[leftmargin=*,topsep=2pt,itemsep=1pt]
% \item $\lambda_{\mathrm{marg}}^{\mathrm{GT}} = 1.0$, $\lambda_{\mathrm{concept}}^{\mathrm{GT}} = 0.5$, $\lambda_{\mathrm{edge}}^{\mathrm{GT}} = 0.5$.
% \item Temperature $\tau_{\mathrm{ic}} = 0.1$.
% \item Margins $\mu_{\mathrm{strict}} = 1.0$, $\mu_{\mathrm{relax}} = 0.3$.
% \item Learning rate $5\times 10^{-6}$ (vs. $5\times 10^{-4}$ in Stage I--2), to ensure calibration rather than overriding.
% \item Adapters updated: same as Stage II (semantic projection MLP, fusion gate, marginal head, cluster Gaussians).
% \end{itemize}

\section{EM-style Optimization}\label{app:em}

This appendix details the E-step / M-step alternation summarized in \S\ref{sec:3.4.1}.

\begin{algorithm}[h]
\caption{Stage II EM-style optimization.}
\label{alg:em}
\small
\begin{algorithmic}[1]
\REQUIRE Pre-trained encoder $f_\theta$ with adapters $\theta_a$; taxonomy backbone $\mathcal{T}^{\mathrm{tax}}$; epochs $T$.
\STATE Initialize $\mathcal{T}^{(0)} \gets \mathcal{T}^{\mathrm{tax}}$.
\FOR{$t = 1, \ldots, T$}
    \STATE \textbf{E-step:}
    \STATE \quad Freeze $\theta_a$; compute $\{\mathbf{e}_i\}$ via current encoder.
    \STATE \quad Run Algorithm~\ref{alg:tree-builder} with temporal admissibility filter $\mathcal{F}_\epsilon$.
    \STATE \quad Obtain $\mathcal{T}^{(t)}, \mathcal{C}^{(t)}$ and identify marginal candidates $\mathcal{M}^{(t)}$.
    \STATE \quad Generate pseudo-labels $\{\tilde{y}_i^{\mathrm{m}}\}$ via Eq.~\ref{eq:marginal-def}.
    \STATE \textbf{M-step:}
    \STATE \quad Freeze $\mathcal{T}^{(t)}$; update $\theta_a$ by one epoch of SGD on $\mathcal{L}^{\mathrm{evo}}$.
\ENDFOR
\RETURN Adapters $\theta_a$, final tree $\mathcal{T} = \mathcal{T}^{(T)}$.
\end{algorithmic}
\end{algorithm}

\paragraph{E-step frequency.} In practice we run the E-step once per epoch. Running it more frequently (e.g., every 100 steps) did not improve performance and increased compute by $\sim 3\times$.

\paragraph{Temporal admissibility filter $\mathcal{F}_\epsilon$.} A candidate merge of clusters $C_p, C_q$ is admissible if for each subsequent merging step the resulting parent does not violate parental temporal ordering by more than $\epsilon = 1$ year, where the nominal time of a cluster is $\tau^{\text{att}}$ (Eq.~\ref{eq:tau-defs}), the mean publication year of its directly attached papers, as in Eq.~\ref{eq:merge-filter}.

\section{Generation-quality Evaluation Protocol}
\label{app:genqual}

\subsection{Evaluation Setup}
Both protocols receive identical supporting information: paper titles, publication years, citation relations, and the anonymized generated hierarchy. Method names, model descriptions, and metric results are withheld. To reduce presentation bias, structures are assigned randomized anonymous identifiers and presented in randomized order. The LLM judge is the same model used for concept labeling, evaluating each structure over multiple independent runs under a fixed prompt and decoding configuration. Human evaluation is conducted by three CS graduate students who did not participate in constructing the FS annotations; they evaluate each structure independently and do not discuss individual cases before submitting their judgments.

\subsection{Criteria and Aggregation}
Both protocols score four dimensions---conceptual organization, scientific evolution, transitional-paper placement, and overall explanatory quality---on a 1--10 scale under the same rubric, each accompanied by a brief evidence-based justification. Scores are averaged over evaluation runs (LLM) or evaluators (human), then over the FS domains, and we report the mean and standard deviation per method and dimension in Table~\ref{tab:genqual}.

The following prompt is given to the LLM judge; human evaluators receive the same criteria in written form.

\begin{tcolorbox}
[colback=gray!5, colframe=gray!50, title=Generation-quality evaluation prompt, fonttitle=\small\bfseries, fontupper=\small\itshape, width=\columnwidth,]
You are an expert researcher. Given a citation graph and a generated hierarchy, evaluate the hierarchy on the following aspects.

\medskip
Citation graph: \texttt{[titles, publication years, citation relations]}

\medskip
Generated hierarchy: \texttt{[anonymized structure]}

\medskip
Aspects:
\begin{enumerate}
    \item \textbf{Conceptual Organization} --- Are papers grouped into semantically coherent concepts? Are concept names accurate?
    \item \textbf{Scientific Evolution} --- Does the hierarchy reflect the historical development of the field? Are parent--child relationships reasonable?
    \item \textbf{Transitional Papers} --- Are transitional / bridge papers attached at an appropriate position?
    \item \textbf{Overall Quality} --- How well does the hierarchy explain the organization and evolution of this research field?
\end{enumerate}
Give a score from 1--10 for each aspect and provide a brief explanation.
\end{tcolorbox}

\end{document}